\documentclass[sn-mathphys-num]{sn-jnl}

\usepackage{graphicx}%
\usepackage{multirow}%
\usepackage{amsmath,amssymb,amsfonts}%
\usepackage{amsthm}%
\usepackage{mathrsfs}%
\usepackage[title]{appendix}%
\usepackage{xcolor}%
\usepackage{textcomp}%
\usepackage{manyfoot}%
\usepackage{booktabs}%
\usepackage{algorithm}%
\usepackage{algorithmicx}%
\usepackage{algpseudocode}%
\usepackage{listings}%

\usepackage{multirow}%
\usepackage{longtable}%
\usepackage{caption}%
\usepackage{subfig}%
\usepackage{url}%
\usepackage{booktabs}

\usepackage{pifont} 
\newcommand{\cmark}{\ding{51}}  
\newcommand{\xmark}{\ding{55}} 
\newcommand{\hmark}{$\approx$}

\theoremstyle{thmstyleone}%
\theoremstyle{thmstyletwo}%

\theoremstyle{thmstylethree}%

\begin{document}

\title[A Survey on Group Fairness in Federated Learning]{A Survey on Group Fairness in Federated Learning: Challenges, Taxonomy of Solutions and Directions for Future Research}


\author[1]{\fnm{Teresa} \sur{Salazar}}\email{tmsalazar@dei.uc.pt (Corresponding author)}
\author[2]{\fnm{Helder} \sur{Araujo}}\email{helder@isr.uc.pt}
\author[3]{\fnm{Alberto} \sur{Cano}}\email{acano@vcu.edu}
\author[1]{\fnm{Pedro} \sur{Henriques Abreu}}\email{pha@dei.uc.pt}

\affil[1]{\orgname{Centre for Informatics and Systems}, \orgdiv{Department of Informatics Engineering of the University of Coimbra}, \orgaddress{\city{Coimbra}, \postcode{3030-290}, \country{Portugal}}}

\affil[2]{\orgname{Institute of Systems and Robotics}, \orgdiv{Department of Electrical and Computer Engineering of the University of Coimbra (ISR) PT}, \orgaddress{\city{Coimbra}, \postcode{3030-290}, \country{Portugal}}}

\affil[3]{\orgname{Virginia Commonwealth University}, \orgaddress{\city{Richmond, VA}, \postcode{23284}, \country{USA}}}


\abstract{Group fairness in machine learning is an important area of research focused on achieving equitable outcomes across different groups defined by sensitive attributes such as race or gender. Federated Learning, a decentralized approach to training machine learning models across multiple clients, amplifies the need for fairness methodologies due to its inherent heterogeneous data distributions that can exacerbate biases. The intersection of Federated Learning and group fairness has attracted significant interest, with 48 research works specifically dedicated to addressing this issue. However, no comprehensive survey has specifically focused on group fairness in Federated Learning. In this work, we analyze the key challenges of this topic, propose practices for its identification and benchmarking, and create a novel taxonomy based on criteria such as data partitioning, location, and strategy. Furthermore, we analyze broader concerns, review how different approaches handle the complexities of various sensitive attributes, examine common datasets and applications, and discuss the ethical, legal, and policy implications of group fairness in FL. We conclude by highlighting key areas for future research, emphasizing the need for more methods to address the complexities of achieving group fairness in federated systems.}

\keywords{Fairness, Bias Mitigation, Federated Learning, Distributed Machine Learning}



\clearpage\maketitle
\thispagestyle{empty}

\section{Introduction}

Group fairness in machine learning refers to the principle that predictions should not prejudice unprivileged groups of the population with respect to sensitive attributes such as race or gender \cite{mehrabi2021survey, pessach2022review, caton2020fairness, survey-infovis}. Ensuring group fairness is essential to prevent discrimination in automated decision-making processes. However, achieving group fairness is a challenging task due to the inherent biases present in data, the difficulty of balancing fairness with other objectives, and the complexity of handling intersectional group identities.

Federated Learning (FL) is an emerging paradigm in machine learning that allows multiple clients to collaboratively train a model while keeping their data decentralized \cite{Federated_Learning_Avg}. This approach is particularly beneficial for preserving privacy, as data remains on local devices rather than being centralized in one location. However, the decentralized nature of FL introduces additional challenges in achieving group fairness. The heterogeneity of data across clients and the limited visibility into the overall data distribution with respect to sensitive attributes challenge the implementation of fairness-aware algorithms in FL settings.

There has been a growing interest in ensuring group fairness in FL, with numerous studies proposing various techniques to address fairness issues. Despite this increased focus, there is no dedicated comprehensive survey that specifically addresses group fairness in the context of FL. This survey aims to fill this gap by providing a comprehensive overview of the challenges, solutions, broader concerns and future directions for achieving group fairness in FL.

To comprehensively review the existing literature on this area of research, we employed a structured search using Google Scholar, where we designed a query to capture a broad range of articles focused on fairness within the context of FL. This query was designed to retrieve all articles that contain the word `federated' in the title and include any of the terms `fair', `fairness', `bias', `equitable', or `equal' in the title. This search yielded a total of 231 research works. We selected Google Scholar for its extensive citation network, broad coverage, and support for boolean operators in search queries, encompassing both peer-reviewed and non-peer-reviewed literature.

The collected research works were systematically categorized based on their nature, resulting in the following categories: articles (216 works), surveys (eight works discussed in Section \ref{section:related-work}), tutorials (one publication), project proposals (four works), and MSc or PhD thesis (two works). We then applied specific inclusion criteria to identify research that explicitly addresses group fairness in FL. A work was considered `specifically dedicated' if it proposed or analyzed methods, metrics, or evaluations targeting group fairness. Based on these criteria, we identified 48 articles published up to March 2025 that were specifically dedicated to group fairness in FL, with the earliest work dating back to 2020. The remaining research works addressed other types of fairness, as discussed in Section \ref{section:types}.

\paragraph*{\textbf{Contributions}} This paper provides a comprehensive survey of group fairness in FL by categorizing and analyzing existing approaches while highlighting key challenges and identifying future research directions. Our main contributions are as follows: 
\begin{itemize}
    \item \textit{Overview of Challenges:} We detail the unique challenges of achieving group fairness in FL, such as the complexities involved in preserving client privacy concerning sensitive attributes and managing heterogeneous data distributions. These factors make achieving group fairness in FL significantly more challenging than in traditional centralized learning systems. 
    \item \textit{Best Practices for Identifying and Benchmarking:} We discuss strategies for identifying group bias in FL and provides a guidelines for benchmarking fairness, addressing important aspects such as heterogeneity, system constraints, and trade-offs.
    \item \textit{Development of a Taxonomy of Approaches:}  We develop the first taxonomy of group fairness approaches in FL, structured around six dimensions: (1) Data Partitioning: how data is partitioned among clients; (2) Location: where the fairness mechanism is implemented; (3) Strategies: specific techniques employed to achieve group fairness; (4) Concerns: broader issues associated with achieving group fairness in FL; (5) Sensitive Attributes: how different approaches manage sensitive groups and their intersections to ensure equitable outcomes; (6) Datasets and Applications: the datasets and application domains commonly used in fair FL studies.
    \item \textit{Ethical, Legal, and Policy Considerations:} We explore the broader ethical and regulatory implications of achieving group fairness in FL. This includes compliance with key legal frameworks, as well as ethical concerns related to bias mitigation, transparency, and accountability in decentralized learning systems. 
    \item \textit{Identification of Research Gaps:} We identify gaps in the existing literature, analyzing areas that warrant further investigation, such as managing intersectionality, developing frameworks for studying group fairness in FL, and addressing challenges in less explored areas.
\end{itemize}

The remainder of this work is structured as follows: Section \ref{section:related-work} discusses the related surveys, Section \ref{section:background} provides the background on group fairness and FL, Section \ref{section:challenges} discusses the challenges of achieving group fairness in FL, Section \ref{section:identifying-benchmarking} discusses the best practices for identifying and benchmarking group fairness in FL, Sections \ref{section:data-partition}, \ref{section:location}, \ref{sec:strategies}, \ref{section:concerns}, \ref{section:sensitive-attributes} and \ref{section:datasets-and-applications} discuss the current works based on data partition, location, strategies, concerns, sensitive attributes, and datasets and applications, Section \ref{section:ethical-legal-policy} presents ethical, legal, and policy considerations of achieving group fairness in FL, Section \ref{section:future-directions} explores future directions for research in this area, and Section \ref{section:conclusions} presents the conclusions of this work.

\section{Related Work}\label{section:related-work}

We review existing surveys and research on fairness in FL, highlighting the gap in detailed coverage of group fairness. While several existing works discuss fairness in FL, we find that they either provide a general overview of different fairness types (as discussed in Section \ref{section:types}) or focus on specific domains, leaving a gap in the comprehensive analysis of group fairness in FL.

Table \ref{tab:related-work} presents a comparison of the related surveys on group fairness in FL. Our analysis is structured according to the key dimensions explored in this work: challenges, identifying and benchmarking fairness, location of fairness interventions, data partitioning, strategies, sensitive attributes, datasets and applications, and ethical, legal, and policy considerations. 

\begin{table}[h]
\centering
\begin{tabular}{c|c|c|c|c|c|c|c|c|c}
\toprule
\multirow{2}{*}{\textbf{Work}} & \multicolumn{9}{c}{\textbf{Group Fairness in FL}} \\ \cmidrule{2-10}
& Challenges & I.B. & D.P. & LOC & Strategies & Concerns & S.A. & D.A. & E.L.P. \\ \toprule
\cite{chen2023privacy} & \cmark & \xmark & \xmark & \xmark & \hmark & \hmark & \xmark & \xmark & \xmark \\ \midrule
\cite{rafi2024fairness} & \cmark & \xmark & \hmark & \hmark & \hmark & \hmark & \xmark & \xmark & \xmark \\ \midrule
\cite{huang2023federated} & \xmark & \xmark & \xmark & \xmark & \xmark & \xmark & \xmark & \xmark & \xmark \\ \midrule
\cite{shi2023towards} & \xmark & \xmark & \xmark & \hmark & \hmark & \xmark & \xmark & \xmark & \xmark \\ \midrule
\cite{vucinich2023current} & \cmark & \xmark & \hmark & \xmark & \hmark & \hmark & \xmark & \xmark & \xmark \\ \midrule
\cite{mashhadi2022fairness} & \hmark & \xmark & \xmark & \xmark & \hmark & \xmark & \xmark & \xmark & \xmark \\ \midrule
\cite{annapareddy2023fairness} & \hmark & \xmark & \xmark & \xmark & \hmark & \xmark & \xmark & \xmark & \xmark \\ \midrule
\cite{ude2023survey} & \cmark & \xmark & \xmark & \xmark & \hmark & \xmark & \xmark & \xmark & \xmark \\ \midrule
\textbf{Ours} & \cmark & \cmark & \cmark & \cmark & \cmark & \cmark & \cmark & \cmark & \cmark \\
\bottomrule
\end{tabular}
\caption{
Comparison to related surveys on group fairness in Federated Learning. While some works may analyze other types of fairness as discussed in Section \ref{section:types}, this table specifically focuses on group fairness. \\
Identifying and Benchmarking (I.B.), Data Partition (D.P.), Location (LOC), Sensitive Attributes (S.A.), Datasets and Applications (D.A.), Ethical, Legal and Policy considerations (E.L.P.). \\
\cmark\ – fully addresses it, \xmark\ – does not address it, \hmark\ – partially addresses it.
}
\label{tab:related-work}
\end{table}

Chen \textit{et al.} \cite{chen2023privacy} provide a survey that addresses privacy and fairness, exploring the trade-offs between them. Loooking at Table \ref{tab:related-work}, it can be observed that they cover fairness challenges but do not address identifying and benchmarking fairness, data partitioning, or intervention location. Their discussion of fairness strategies and concerns remains partial, and they omit a discussion on sensitive attributes considerations, datasets and applications, and ethical, legal and policy considerations. In contrast to Chen \textit{et al.}, whose primary focus lies in the tension between privacy and fairness, our survey isolates group fairness as a distinct axis of study and presents a novel taxonomy for categorizing different works. Rafi \textit{et al.} \cite{rafi2024fairness} also examine the intersection of privacy and fairness in FL. They discuss fairness challenges and provide partial coverage of data partitioning, intervention location, strategies, and concerns. However, they too omit identifying and benchmarking fairness, sensitive attributes, datasets and applications, and ethical, legal, and policy aspects. Huang \textit{et al.} \cite{huang2023federated} focus on generalization, robustness, and fairness, but their notion of fairness is limited to collaboration and performance fairness. They do not address any aspects of group fairness, leaving all key dimensions uncovered. Shi \textit{et al.} \cite{shi2023towards} and Vucinich \textit{et al.} \cite{vucinich2023current} review various fairness notions in FL. Shi \textit{et al.} provide partial coverage of intervention location and strategies, but do not address other dimensions. Vucinich \textit{et al.} discuss fairness challenges and partially address data partitioning, strategies, and concerns, yet similarly overlook identifying and benchmarking fairness, sensitive attributes, datasets and applications, and ethical, legal, and policy issues. Mashhadi \textit{et al.} \cite{mashhadi2022fairness} narrow their scope to spatio-temporal applications. While they partially address fairness challenges and propose specific metrics, their coverage of strategies is limited and they do not consider broader issues such as sensitive attributes or applications outside their domain. Annapareddy \textit{et al.} \cite{annapareddy2023fairness} restrict their discussion to healthcare, partially covering fairness challenges and strategies but ignoring the wider FL landscape. Finally, Ude \textit{et al.} \cite{ude2023survey} explicitly target group fairness, but their treatment is outdated and limited. They cover fairness challenges and provide a partial discussion of strategies, yet omit all other dimensions, including benchmarking, sensitive attributes, and policy implications. Across all these surveys, sensitive attributes - which are important to defining fairness constraints - are entirely overlooked. Similarly, datasets and applications related to fairness in FL are absent from existing discussions. Finally, ethical, legal, and policy considerations, which are important for real-world deployment, are also not covered by any prior survey.

In contrast, our work provides the first survey that systematically covers all of the dimensions presented in Table \ref{tab:related-work}. We address fairness challenges, introduce mechanisms for identifying and benchmarking group fairness, categorize strategies by data partitioning and intervention location, and extend the discussion to sensitive attributes, datasets, applications, and ethical, legal, and policy concerns. While some prior works address multiple types of fairness (as discussed in Section \ref{section:types}), they do not explore the specifics of group fairness. Yet, the mechanisms for ensuring group fairness are inherently different from those used to achieve other fairness notions. Our comprehensive treatment therefore fills an important gap in the literature, providing both theoretical insights and practical guidance for researchers and practitioners, and underscoring the need for a dedicated survey on group fairness in FL.

Finally, it is also important to position our work against broader survey areas that intersect with our topic, including fairness in machine learning \cite{mehrabi2021survey, pessach2022review, caton2020fairness}, privacy-preserving FL \cite{mothukuri2021survey, yin2021comprehensive}, and trustworthy AI \cite{li2023trustworthy, kowald2024establishing}. Surveys on fairness in machine learning \cite{mehrabi2021survey, pessach2022review, caton2020fairness} provide rich discussions of group fairness definitions and metrics, but they do not address the unique challenges introduced by the federated setting, such as non-IID data distributions and communication constraints. Privacy-preserving FL surveys \cite{mothukuri2021survey, yin2021comprehensive} primarily emphasize security and privacy trade-offs, often overlooking fairness concerns or treating them only tangentially. Trustworthy AI surveys \cite{li2023trustworthy, kowald2024establishing}, meanwhile, cover fairness as one of several dimensions (alongside robustness, explainability, and privacy), but they do not provide a systematic treatment of group fairness in FL. Our survey is therefore complementary to these broader perspectives, while being the first to provide a focused and comprehensive synthesis of group fairness in FL settings.

\section{Background}\label{section:background}

In this section, we provide the necessary background on FL and fairness, which is essential for understanding the remainder of this survey.

\subsection{Federated Learning}

FL is a decentralized approach to training machine learning models that enables multiple devices or organizations to collaborate without sharing their raw data \cite{Federated_Learning_Avg, flsurvey}. This paradigm was introduced to address privacy concerns, data security issues and high communication costs associated with traditional centralized machine learning methods, where data from various sources is aggregated in a single location for training.

In FL, the model training process is decentralized. Each participating device, referred to as a client, downloads a global model from a central server. The client then trains the model locally using its own data and subsequently sends only the updated model updates back to the server. The central server aggregates these updates from all clients to improve the global model. This iterative process continues until the model converges.

Federated Averaging (FedAvg) \cite{Federated_Learning_Avg} was the first FL algorithm to be introduced, in which the central server computes a weighted average of the updates. At its core, FedAvg operates through a series of coordinated steps between a central server and multiple participating clients.

\begin{figure}[h!]
    \centering
    \includegraphics[width=0.8 \linewidth]{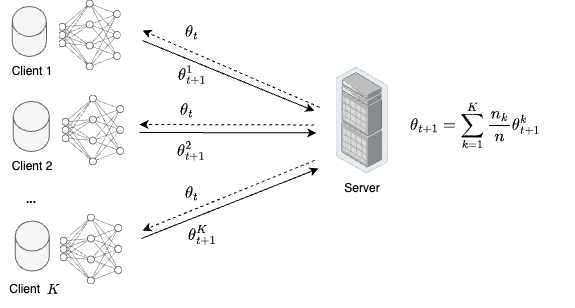}
    \caption{Illustration of the Federated Averaging algorithm with $K$ clients participating in the federation.}
    \label{fig:fedavg-diagram}
\end{figure}

Figure \ref{fig:fedavg-diagram} presents a diagram demonstrating the FedAvg algorithm with $K$ clients participating in the federation. Initially, the server initializes the global model, denoted as $\theta_0$. Each client selected for participation at each communication round $t$ performs a local update to refine the global model based on its own local dataset. This local update is achieved by minimizing a local loss function $F_k(\theta_t)$ using gradient descent or other optimization algorithm, where $\theta_t$ represents the current global model at communication round $t$, and $\eta$ denotes the learning rate. The updated model for client $k$ is computed as $\theta_{t+1}^{k} = \theta_t - \eta \nabla F_k(\theta_t)$.

Subsequently, the server aggregates these updated models from all participating clients to generate an improved global model for the next communication round. The aggregation process involves computing a weighted average of the local updates, where the weights are proportional to the number of datapoints $n_k$ held by each client $k$. Formally, the aggregated model $\theta_{t+1}$ at communication $t+1$ is computed as:
\begin{equation}
    \theta_{t+1} = \sum_{k=1}^K \frac{n_k}{n} \theta_{t+1}^k ,
\end{equation}
where $n$ is the total number of datapoints across all clients. This process is repeated for several communication rounds.

In terms of the overall objective, the goal of the FedAvg algorithm is to minimize a global objective function $F(\theta)$, which is defined as the weighted sum of local objective functions across all participating clients. Formally, the global objective function is given by:

\begin{equation}
    \min_{\theta} F(\theta) = \sum_{k=1}^{K} \frac{n_{k}}{n} F_k(\theta)
\label{eq:min}
\end{equation}
where $K$ is the total number of clients, $n_k$ is the number of datapoints of client $k$, $n = \sum_{k=1}^K n_k$ is the total number of datapoints across all clients, and $F_k(\theta)$ is the local objective function for client $k$ \cite{Federated_Learning_Avg}.

The local objective function $F_k(\theta)$ for client $k$ can be written as:
\begin{equation}
    F_k(\theta) = \frac{1}{n_k} \sum_{i \in \mathcal{D}_k} f_i(\theta) ,
\end{equation}
where $\mathcal{D}_k$ is the local dataset of client $k$, and $f_i(\theta)$ is the loss function for the $i$-th datapoint in the local dataset.

\subsubsection{Advantages}

FL offers substantial advantages that make it a compelling approach for training machine learning models across distributed environments.

\paragraph{Enhanced Privacy and Security} 
By keeping data on local devices and only sharing model updates, this decentralized approach ensures that sensitive information remains under the control of individual clients, thereby enhancing data privacy \cite{flsurvey, saha2024multifaceted}. Moreover, FL frameworks often incorporate encryption and differential privacy techniques to further safeguard data during the aggregation process \cite{el2022differential}.

\paragraph{Reduced Latency} Local processing can lead to faster model training, which is specially useful in applications where timely responses are critical \cite{li2020federated}. This reduction in latency is particularly beneficial in edge computing scenarios, where data processing occurs closer to the source, minimizing delays associated with data transmission to a centralized server.

\paragraph{Flexible Scalability} With the growing capabilities of edge devices and the increasing volume of individual data, centralizing all data to a single server can result in under-utilization of edge computing power. FL's distributed framework enables efficient utilization of the computational resources available across numerous devices spread across various geographical locations \cite{zhang2022federated}. This approach, therefore, overcomes scalability challenges by parallelizing the computation across multiple devices.

\paragraph{Regulatory Compliance} FL aligns well with data protection regulations by ensuring that sensitive data does not leave its original location. This compliance with regulatory frameworks such as the General Data Protection Regulation (GDPR) \cite{gdpr} in Europe and HIPAA (Health Insurance Portability and Accountability Act) \cite{act1996health} in the United States is important for organizations handling sensitive data.

\subsubsection{Challenges}\label{section:fl-challenges}

Despite its promising benefits, FL encounters several significant challenges that should be addressed to achieve its full potential across various applications.

\paragraph{Communication Overhead} 
Frequent transmission of model updates between local devices and the central server can lead to significant communication costs. This overhead arises from the need to synchronize and aggregate updates from multiple clients, especially in large-scale FL setups. Efficient communication protocols can be used to mitigate these costs and ensure the scalability of FL \cite{chen2021communication}.

\paragraph{Data Heterogeneity} 
Data across clients can be non-IID (not Independent and Identically Distributed), leading to challenges in model convergence and performance \cite{flsurvey}. Non-IID means that each client's dataset may not follow the same underlying distribution, and the datapoints within a client's dataset may not be independent of each other. For example, one client's data could be heavily skewed toward certain classes (e.g., only images of dogs), while another client may have data biased toward entirely different classes (e.g., only images of cats). This lack of uniformity contrasts with the IID assumption in centralized learning, where data is assumed to be drawn independently from the same distribution for all clients. Addressing data heterogeneity requires adaptive algorithms that can effectively aggregate diverse data sources while preserving performance across the federated network.

\paragraph{System Heterogeneity} 
Clients may have varying computational capabilities, network connectivity, and energy resources, complicating the coordination of the FL training process. This system heterogeneity introduces challenges in resource allocation and workload management across FL systems \cite{ye2023heterogeneous, shanmugarasa2023systematic}. Adaptive scheduling algorithms and resource-aware optimization strategies can be used to ensure equitable participation and efficient utilization of client resources.

\paragraph{Privacy and Security Risks} 
While FL removes the need for direct data sharing, it remains vulnerable to several privacy and security threats \cite{lyu2020threats, hu2024overview}. Model inference attacks, for instance, can occur when adversaries deduce sensitive information from the shared model updates. Similarly, poisoning attacks involve adversaries deliberately introducing corrupted data or malicious updates to skew the model's performance. Protecting against these threats requires robust defensive strategies to ensure the integrity and confidentiality of the FL process.

\subsection{Types of Fairness in Federated Learning}\label{section:types}

Ensuring fairness in FL is important due to the diverse and heterogeneous nature of both the data and the participants. Several types of fairness have been identified in FL, each addressing different aspects of fairness.

\paragraph{Group fairness} Also known as demographic fairness, this principle promotes equity in the outcomes of machine learning models across protected and unprotected groups defined by sensitive attributes such as race, gender, or age \cite{mehrabi2021survey, pessach2022review, caton2020fairness, survey-infovis}. For example, if a model systematically favors male applicants over female applicants for a job recommendation task, it violates group fairness by producing biased outcomes based on gender. In the context of FL, each client may hold data from multiple sensitive groups, meaning it cannot be assumed that each client belongs to only one sensitive group. Group fairness is the focus of this work and more details on group fairness are presented in the next sections.

\paragraph{Individual fairness} This notion requires that similar individuals receive similar outcomes from the machine learning model \cite{dwork2012fairness}. This means that the model should treat individuals with similar characteristics in a similar manner. In FL, this could mean that two patients with similar medical histories across different hospitals (i.e. different clients) should receive similar predictions in a federated healthcare model, regardless of which hospital (client) their data belongs to. Ensuring individual fairness can be challenging in FL because clients may have non-overlapping feature distributions, making it difficult to measure similarity across different datasets.

\paragraph{Performance distribution fairness} Also known as client fairness, this principle requires that the performance of the FL model, such as accuracy, is evenly distributed across all clients \cite{li2019fair, gao2024does, chaudhury2024fair}. This concept emphasizes the importance of uniformity in performance, ensuring that no single client is disproportionately advantaged or disadvantaged. For example, in a federated loan approval system where each client is a specific bank, if the model performs well for larger banks but poorly for smaller banks, it creates unfair disparities in model accuracy.

\paragraph{Selection fairness} Also known as client participation fairness, this type focuses on the fairness in selecting clients to participate in the FL communication rounds \cite{cho2022towards, sun2025debiasing}. In each round of FL, a subset of clients is selected to update the global model. Selection fairness ensures that this process is unbiased and that all clients have an equitable opportunity to participate. This is important to prevent biases that could arise from consistently selecting certain clients over others. For example, in a mobile keyboard FL system, consistently selecting high-end devices for training while excluding low-end ones can lead to biased predictions.

\paragraph{Contribution fairness} Also known as collaborative fairness, this principle is concerned with providing appropriate incentives for clients to participate in the FL process \cite{zeng2021comprehensive, fan2024fair, pan2024towards, wang2024fedsac}. It ensures that a client's reward is proportional to its contribution to the global model. This is important for motivating clients to actively participate and contribute with high-quality data. For example, in a federated traffic prediction model, if a city contributing with high-quality data receives the same reward as one providing noisy data, it discourages fair collaboration.

Each type of fairness in FL addresses different aspects of equity and justice in the model training process. While this work primarily focuses on group fairness, it is important to understand and differentiate these types of fairness from each other. For the sake of simplicity, in the remainder of this survey, the term `fairness' refers specifically to group fairness.

\subsection{Group Fairness in Machine Learning}

Group fairness in machine learning aims to ensure that algorithmic decisions do not disproportionately benefit or harm specific demographic groups. This involves considering sensitive attributes, which are characteristics of individuals that, when used in decision-making processes, could lead to discriminatory outcomes \cite{mehrabi2021survey, pessach2022review, caton2020fairness, survey-infovis}. Common sensitive attributes include race, gender, age, and socioeconomic status.

In the context of group fairness, individuals can be categorized into protected and unprotected groups based on their sensitive attributes. Protected groups are groups of individuals who belong to categories that have historically been disadvantaged or subject to discrimination. For example, in the context of hiring practices, women might be considered a protected group if they have been historically under-represented in certain industries, such as technology or engineering. Similarly, in the context of lending or credit approval, individuals from Black or Hispanic communities, may be considered protected groups due to historical discrimination in access to financial services. On the other hand White men individuals would be considered an unprotected group in these scenarios.

\subsubsection{Metrics}

Many statistical measures of group fairness in binary classification rely on metrics that can be explained using a confusion matrix that is often used to describe the performance of a classification model \cite{fairness_definitions_explained, review_fair_ml, comparative_study_fairness_interventions}. Here, $S$ represents the sensitive attribute with two groups ($S=0$ and $S=1$), $Y$ represents a target class where $1$ is the positive class and $0$ is the negative class (e.g. receiving a loan or not), and $\hat{Y}$ is the predicted class. In a confusion matrix the rows and columns represent instances of the predicted and actual classes, respectively. The confusion matrix is presented in Table \ref{tab:conf-matrix}.

Looking at the confusion matrix, one can derive a measure of the ratio (RAT) or the difference (DIF) of True Positive Rates (TPR) between two groups, also known as Equality of Opportunity \cite{hardt2016-equality-of-opportunity}:

\begin{equation}
    \begin{gathered}
        \frac{P [\hat{Y} = 1 \mid S = 0, Y = 1]}{P [\hat{Y} = 1 \mid S = 1, Y = 1]} \quad OR 
        \quad P[\hat{Y} = 1 \mid S = 0, Y = 1] - P [\hat{Y} = 1 \mid S = 1, Y = 1]
    \end{gathered}
\end{equation}

When using the ratio or the difference for a specific metric, values of 1 and 0 indicate the best fairness results, respectively. Additionally, in contexts with multiple groups, it is also common to access fairness by reporting group-specific metrics individually for each group (GS), calculating the average across all groups (AVG), or analyzing disparities using standard deviation-based (STD) or variance-based (VAR) metrics.

\begin{table}[h]
\centering
\begin{tabular}{c|c|c}
\toprule
& \shortstack[c]{Actual Positive \\ $Y = 1$} & \shortstack[c]{Actual Negative \\ $Y = 0$}  \\
\toprule
\multirow{3}{*}{\shortstack[c]{Predicted \\ Positive \\ $\hat{Y} = 1$}} & True Positive (TP) & False Positive (FP) \\
& $TPR/Recall = P(\hat{Y}= 1 | Y = 1) = \frac{TP}{TP + FN}$ 
& $FPR = P(\hat{Y}= 1 | Y = 0) = \frac{FP}{FP + TN}$ \\
& $PPV/Precision = P(Y = 1 | \hat{Y} = 1) = \frac{TP}{TP + FP}$ 
& $FDR = P(Y = 0 | \hat{Y} = 1) = \frac{FP}{FP + TP}$ \\
\midrule
\multirow{3}{*}{\shortstack[c]{Predicted \\ Negative \\ $\hat{Y} = 0$}} & False Negative (FN) & True Negative (TN) \\
& $FNR = P(\hat{Y}= 0 | Y = 1) = \frac{FN}{FN + TP}$ 
& $TNR = P(\hat{Y}= 0 | Y = 0) = \frac{TN}{TN + FP}$ \\
& $FOR = P(Y = 1 | \hat{Y}= 0) = \frac{FN}{FN + TN}$ 
& $NPV = P(Y = 0 | \hat{Y} = 0) = \frac{TN}{TN + FN}$ \\
\bottomrule
\end{tabular}
\caption{Confusion Matrix.}
\label{tab:conf-matrix}
\end{table}

Fairness metrics can be divided into five groups: metrics conditioned of the outcome, metrics conditioned on the decision, performance-based metrics, unconditional metrics, and loss-based metrics.

\paragraph{Conditioned on the Outcome} 
The definitions of fairness conditioned on the outcome, $Y$, can be divided into two groups. The first group is conditioned on $Y = 0$, and demands Equality of False Positive Rates (FPR) (also known as Predictive Equality) or Equality of True Negative Rates (TNR) between two sensitive groups. These types of metrics can be considered, for example, from the perspective of innocent defendants by requiring that individuals who do not go on to be re-arrested have the same probability of being released regardless of their sensitive attribute value.

On the other hand, the second group is conditioned on $Y = 1$, and demands Equality of True Positive Rates (TPR) (also known as Equality of Opportunity \cite{hardt2016-equality-of-opportunity}) or Equality of False Negative Rates (FNR) between two sensitive groups. These types of metrics can be considered, for example, from the perspective of people that apply to receive a loan to have the same likelihood of receiving a loan, regardless of whether they belong to the protected or unprotected group.

The types of metrics conditioned on the outcome are more aligned with the perspective of the population evaluated by the model as they demand that individuals who are similar with respect to their outcomes be treated similarly \cite{catalogue-metrics}.

\paragraph{Conditioned on the Decision}

The definitions of fairness conditioned on the decision, $\hat{Y}$, can be divided into two groups. The first group is conditioned on $\hat{Y} = 0$, and demands Equality of False Omission Rates (FOR) or Equality of Negative Predictive Values (NPV) between two sensitive groups. These types of metrics can be considered, for example, for requiring that individuals who were granted a loan to have the same probability to default, regardless of whether they belong to the protected or unprotected group.

On the other hand, the second group is conditioned on $\hat{Y} = 1$, and demands Equality of Positive Predictive Values (PPV) (also known as Predictive Parity \cite{bias-recidivism-prediction}) or Equality of False Discovery Rates (FDR) between two sensitive groups. These types of metrics can be considered, for example, for requiring that people who were classified as criminals to have the same probability of being a criminal, regardless of their sensitive attribute value.

The types of metrics conditioned on the decision reflect fairness in a way that individuals with the same decision would have had similar outcomes, regardless of whether they belonged to the protect or unprotected group \cite{catalogue-metrics}.

\paragraph{Performance-based}

Performance-based fairness metrics are derived from the confusion matrix but are not conditioned on solely the outcome ($Y$) or the decision ($\hat{Y}$). For instance, Overall Accuracy Equality \cite{fairness-criminal-justice} is achieved when the prediction accuracy is equal across groups, meaning that the probability of correctly classifying an individual (whether they belong to the positive or negative class) is the same for all sensitive groups. Another metric is F1-score Equality, which requires the F1-score, a balance between precision and recall, to be equal across groups.

These performance-based metrics assess fairness by ensuring that the model's overall predictive performance does not disproportionately favor any particular group.

\paragraph{Loss-based}

Loss-based fairness metrics focus on ensuring similar losses with respect to a loss function for both protected and unprotected groups. These metrics are commonly used during the training phase of machine learning models to actively guide the learning process toward fair outcomes. Although not as common, researchers also report these metrics during the evaluation phase.

\paragraph{Unconditional}

Unconditional fairness metrics, such as Statistical Parity (SP), are also not conditioned on either the outcome ($Y$) or the decision ($\hat{Y}$). Instead, they evaluate fairness by comparing the overall rates of positive outcomes between protected and unprotected groups. Statistical Parity, for example, requires that the proportion of individuals receiving a positive decision (e.g., being hired, receiving a loan) is equal across groups, regardless of their underlying qualifications or outcomes \cite{mehrabi2021survey}.

For instance, in a hiring context, Statistical Parity would demand that the proportion of hires from a protected group be the same as that from an unprotected group, without factoring in their specific qualifications.

Unconditional fairness metrics are sometimes criticized for ignoring individual merit, but they are valuable in contexts where the goal is to ensure equitable representation or mitigate systemic biases in decision-making processes.

Table \ref{tab:metrics} presents a summary of the most commonly used group fairness metrics.

\begin{table}[h!]
\centering
\begin{tabular}{l|l}
\toprule
\textbf{Group Fairness Metric} & \textbf{Formulation} \\
\toprule
Statistical Parity & $P [\hat{Y} = 1 \mid S = 0] = P [\hat{Y} = 1 \mid S = 1]$ \\ \midrule
Equality of Opportunity & $P [\hat{Y} = 1 \mid S = 0, Y = 1] = P [\hat{Y} = 1 \mid S = 1, Y = 1]$ \\ \midrule
Predictive Equality & $P [\hat{Y} = 1 \mid S = 0, Y = 0] = P [\hat{Y} = 1 \mid S = 1, Y = 0]$ \\ \midrule
Equalized Odds & $P [\hat{Y} = 1 \mid S = 0, Y = y] = P [\hat{Y} = 1 \mid S = 1, Y = y], \quad y \in \{0, 1\}$ \\ \midrule
Predictive Parity & $P [Y = 1 \mid S = 0, \hat{Y} = 1] = P [Y = 1 \mid S = 1, \hat{Y} = 1]$ \\ \midrule
Overall Accuracy Equality & $P [\hat{Y} = Y \mid S = 0] = P [\hat{Y} = Y \mid S = 1]$ \\ 
\bottomrule
\end{tabular}
\caption{Summary of most commonly used group fairness metrics.}
\label{tab:metrics}
\end{table}

\subsubsection{Approaches for Achieving Group Fairness in Machine Learning}

\begin{figure}[h!]
    \centering
    \includegraphics[width=\linewidth]{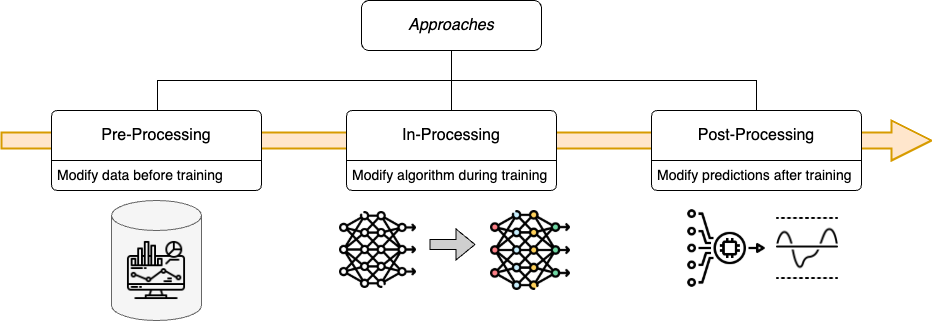}
    \caption{Approaches for achieving group fairness in machine learning.}
    \label{fig:approaches}
\end{figure}

Approaches to achieve group fairness in centralized machine learning are usually categorized into three main types, according to the stage in which they are performed: pre-processing, in-processing, and post-processing \cite{mehrabi2021survey} as illustrated by Figure \ref{fig:approaches}.

\paragraph{Pre-processing Approaches}
Pre-processing approaches aim to mitigate bias before the model training phase. This involves modifying the training data to achieve fairness \cite{salazar2021fawos, kamiran2012data, feng-2019}. Techniques include re-sampling \cite{salazar2021fawos}, where the dataset is adjusted by oversampling under-represented groups or undersampling over-represented groups to balance the data distribution. Relabeling is another technique that involves modifying the labels in the dataset to reduce bias \cite{kamiran2012data}. Another technique is fair representation learning \cite{feng-2019}, which aims to learn new representations of the data that are invariant to sensitive attributes while preserving essential information for prediction tasks.

\paragraph{In-processing Approaches}
In-processing approaches incorporate fairness objectives directly into the model training process \cite{zafar2017fairness, zhang-2018, kamishima2012fairness}. One technique is the inclusion of fairness constraints, where constraints are added to the optimization problem to ensure that the model's predictions satisfy certain fairness criteria \cite{zafar2017fairness}. Adversarial debiasing uses adversarial training to remove bias by training a model that predicts the target variable while an adversary tries to predict the sensitive attribute from the model's predictions \cite{zhang-2018}. Fair regularization involves incorporating regularization terms into the loss function to penalize unfair outcomes \cite{kamishima2012fairness}.

\paragraph{Post-processing Approaches}
Post-processing approaches modify the model's predictions to achieve fairness after the model has been trained \cite{post-1, post-2}. This can involve techniques such as thresholding, where decision thresholds are adjusted for different sensitive groups to equalize outcomes. 

In FL, achieving fairness is more complex due to the involvement of multiple clients and a central server. Approaches can be applied at different locations (more details in Section \ref{section:location}): on the server, at each client, or using a hybrid strategy. The next sections detail the approaches for achieving group fairness in FL.

\section{Challenges}\label{section:challenges}

Achieving group fairness in FL introduces several unique challenges compared to centralized machine learning, primarily due to its decentralized nature and the intrinsic characteristics of federated systems. These challenges are inherent to all fairness-aware FL approaches as each must address them to achieve fairness. Below, we discuss some of the key challenges in developing fairness-aware algorithms in FL, which extend the challenges of FL detailed in Section \ref{section:fl-challenges}.

\paragraph{Data Heterogeneity} In centralized machine learning, training data is often assumed to be IID, meaning that each datapoint is drawn from the same distribution and that the datapoints are statistically independent of one another. However, this assumption does not hold in FL settings, where each client has its own private dataset that may not be representative of the global data distribution \cite{zhao2018federated, zhu2021federated}. This non-IID nature of data across clients can lead to the introduction or exacerbation of biases in the global model. If clients have data that is not representative of the whole population, their contributions to the model could result in biased updates that do not generalize well across all groups. This can negatively impact the performance of the global model, particularly for protected groups, leading to a model that may perform well for some populations while failing to provide equitable outcomes for others \cite{amiri2022impact, abay2021addressing}.

\paragraph{Restricted Information} FL requires training data to be locally stored on clients' devices to protect privacy, which means that the central server cannot access raw training data or sensitive attributes directly \cite{Federated_Learning_Avg}. This restriction limits the ability to apply fairness-aware techniques that rely on global information about the dataset. For instance, in centralized machine learning, algorithms can directly manipulate data or model parameters to mitigate biases by leveraging knowledge about sensitive attributes. In contrast, FL requires innovative methods to ensure fairness without direct access to such detailed information.

\paragraph{Aggregation Algorithms} The aggregation process in FL, where the central server combines model updates from multiple clients, can introduce biases depending on the aggregation strategy used. Common aggregation methods such as FedAvg perform a weighted average of model updates, giving higher importance to updates from clients with more data. This can inadvertently exacerbate biases, particularly if clients with larger datasets do not accurately reflect the broader population, leading to the under-representation of protected groups \cite{salazar2023fair}. Careful design of algorithms is needed to ensure fair representation of all groups.

\paragraph{Limited Client Participation} In FL, not all clients participate in every round of training. This selective participation can lead to biased model updates if certain clients, especially those that contain relevant data from protected groups, are under-represented in the training process \cite{cho2022towards}. Ensuring that clients with diverse data contribute frequently is important for maintaining group fairness across the model's predictions.

\paragraph{Resource Constraints} Clients in FL environments often have varying computational and communication resources. Devices with lower capabilities may struggle to participate fully, potentially skewing the training process towards clients with more resources \cite{yang2021characterizing}. This disparity can create bias in the global model if clients with less resources are systematically excluded from the training rounds. Thus, careful attention to resource constraints is essential to ensure fair contribution and representation of all groups in the FL process.

\paragraph{Long-term Fairness} Ensuring fairness not just in individual training rounds but over the long term is a significant challenge. As models are updated continuously, maintaining fairness over time requires ongoing monitoring and adjustments. This is particularly critical in dynamic environments where client distributions and data characteristics may change \cite{salazar2024unveiling}.

\section{Identifying and Benchmarking}\label{section:identifying-benchmarking}

Ensuring group fairness in FL requires more than proposing fairness-aware methods; it is also important to establish best practices for identifying and benchmarking group fairness in FL systems. The decentralized nature of FL, combined with non-IID data distributions and privacy constraints, makes it more difficult to systematically assess fairness across different demographic groups compared to centralized settings. Without standardized evaluation practices, comparing fairness interventions and understanding their trade-offs with accuracy and privacy remains challenging. Hence, the objective of this section is to outline strategies for detecting group bias in FL and also discuss best practices for benchmarking fair FL methods to ensure practical deployment in real-world applications.

\subsection{Identifying}

Identifying group unfairness in FL begins with defining the groups based on sensitive attributes that are relevant to the specific application of the model, whether it be a single attribute or multiple attributes (more details in Section \ref{section:sensitive-attributes}). 

Once groups are defined, fairness can be assessed. However, since in FL the data is decentralized, traditional methods of fairness evaluation that rely on a shared dataset cannot be directly applied. Hence, there are two main options for identifying unfairness in FL. The first option is to evaluate fairness at the client level using local metrics. Each client can evaluate their model's performance on the relevant sensitive groups and sends aggregated statistics, rather than raw data, to the central server. By aggregating local fairness metrics, the system can get a clearer picture of how the model performs across different groups. 

The second approach to identifying group bias is through the use of a global validation set at the server. In this approach, a validation set consists of data contributed by multiple clients \cite{mehrabi2022towards}. To ensure that the validation set is representative of the overall population, clients can collaborate by contributing samples that reflect the distribution of different sensitive groups. This can be done through stratified sampling, where each client provides data proportional to the demographic composition of its local dataset. Constructing such a validation set presents its own set of challenges, particularly in ensuring the privacy of individual clients while still providing a sufficiently diverse and representative dataset.

Lastly, group bias in FL is not a static issue as it may evolve as the model trains across different communication rounds or even over time \cite{salazar2024unveiling}. Regular monitoring of fairness metrics across these communication rounds or time is important to detecting fairness drift, particularly in real-world FL settings.

\subsection{Benchmarking}

Benchmarking group fairness in FL presents unique challenges due to the decentralized nature of the system, non-IID data, and privacy constraints. To enable consistent comparisons across different fairness-aware FL approaches, it is important to account for different FL settings while maintaining a standardized evaluation process:
\begin{itemize}
    \item \textit{Number of Clients:} the number of participating clients per communication round should be carefully controlled to ensure fair comparisons between different algorithms. A larger number of clients per round may reduce individual client influence, potentially mitigating fairness disparities, whereas a smaller number may amplify biases from specific clients' data distributions.    
    \item \textit{Number of Rounds}: the number of communication rounds should be standardized across experiments to ensure consistent evaluations, and fairness should be measured at different stages of training to detect fairness drift.
    \item \textit{System Heterogeneity and Resource Constraints:} differences in client devices, computational power, and network conditions may introduce biases if lower-resource clients (potentially representing specific sensitive groups) contribute with lower-quality updates, potentially affecting fairness.
    \item \textit{Dataset Heterogeneity:} Standardized data partitioning using non-IID splits should be used to reflect real-world biases while ensuring comparability across experiments. This helps assess how fairness is impacted under different levels of data heterogeneity.
    \item \textit{Model Architectures:} Using identical model architectures across experiments eliminates confounding factors and ensures fairness assessments are not influenced by differences in model design.
    \item \textit{Hyperparameters:} Standardizing learning rates, batch sizes, and optimization strategies minimizes variability, ensuring fairness evaluations are not impacted by tuning inconsistencies.
    \item \textit{Fairness Metrics Selection:} The choice of fairness metrics should be standardized across experiments to enable meaningful comparisons.
\end{itemize}

Finally, it is also important to consider the trade-offs between fairness and other competing objectives. A key trade-off exists between group fairness and model performance, as enforcing fairness constraints may sometimes reduce overall accuracy \cite{mehrabi2021survey}. Similarly, group fairness and privacy must be balanced, as to ensure the fairness clients may need to share more data with the server, which can increase the privacy risk \cite{sun2023toward}. Another important consideration is the distinction between local and global fairness, where fairness improvements at the global level may not translate to equitable outcomes at the local level for specific clients \cite{meerza2024glocalfair}. Lastly, group fairness may also conflict with other fairness notions, such as performance distribution fairness or client selection fairness.

\section{Data Partition}\label{section:data-partition}

FL can be categorized based on how data is partitioned among the clients. The three primary types of data partitioning in FL are horizontal, vertical, and transfer learning \cite{flsurvey}. Achieving group fairness in each of these settings presents unique challenges and requires tailored solutions. In this section, we explain these three types of FL, highlighting the specific challenges and considerations for ensuring group fairness in each context. 

\begin{figure}[h!]
    \centering
    \includegraphics[width=0.925\linewidth]{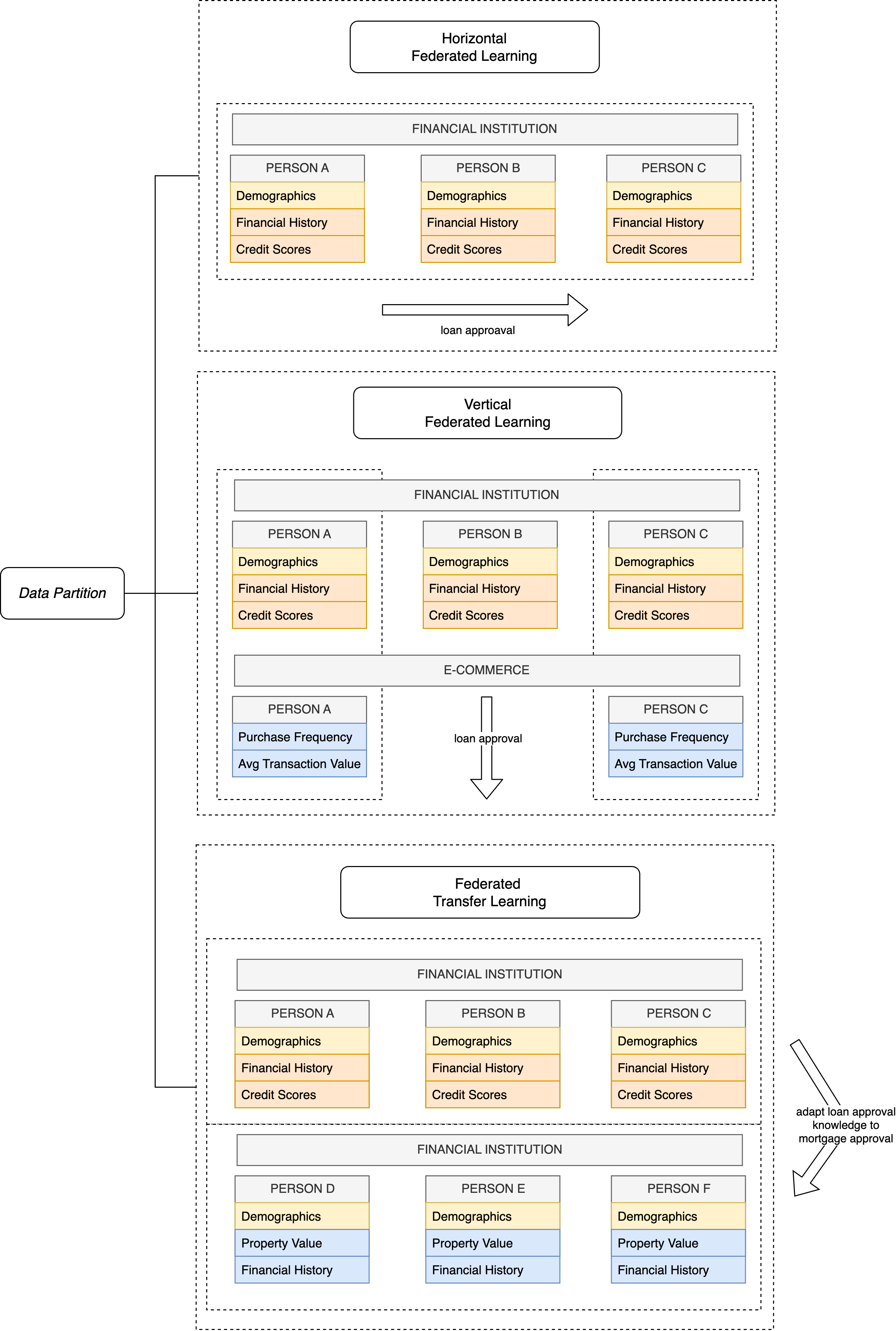}
    \caption{Fair Federated Learning categorized by data partition.}
    \label{fig:data-partition}
\end{figure}

Figure \ref{fig:data-partition} illustrates the three types of FL based on data partitioning. The figure uses a financial institution as a contextual example to demonstrate how different data types (e.g., demographics, financial history, credit scores) are distributed and processed across various FL scenarios. This illustration highlights how each type of FL handles data partitioning in scenarios where the financial institution collaborates with other entities, such as e-commerce platforms, to build models for applications such as loan and mortgage approvals.

\subsection{Horizontal FL}

Most research on group fairness in FL has traditionally focused on Horizontal FL (HFL) (except for \cite{liu2022achieving}). In HFL, clients hold data with the same feature space but different instances \cite{flsurvey}. This approach is particularly relevant in scenarios where multiple organizations or devices have similar types of data for different user groups. The goal is to ensure that the trained model maintains fairness across these diverse datasets without compromising the privacy of individual data sources.

A real-world example of group fairness in HFL could involve financial institutions in different regions collaborating to build a fair predictive model for loan approval. Each bank has the same type of customer data, including demographics, financial history, and credit scores. The FL system would train a model ensuring that the loan approval predictions are fair across different demographic groups, such as age, gender, and ethnicity.

\subsection{Vertical FL}\label{sec:vertical}

The exploration of group fairness in Vertical FL (VFL) is equally important. In VFL, clients hold different subsets of features related to the same group of users \cite{flsurvey, wei2022vertical}. This makes it essential to ensure fairness across these vertical partitions, as each client contributes with unique information to the global model.

A real-world example of group fairness in VFL might involve a collaboration between an e-commerce company and a financial institution. The e-commerce company has data on customers' purchasing behavior such as purchase frequency or average transaction value, while the financial institution has data on customers' demographics, credit scores and financial history. Together, they aim to create a fair model for loan approval. The model must ensure that it does not unfairly discriminate against customers based on sensitive attributes such as race, gender, or socioeconomic status.

However, addressing group fairness in VFL poses additional challenges due to its intrinsic characteristics. Firstly, ensuring the privacy of data across all participating organizations often conflicts with the need for a unified training dataset, which is essential for implementing fairness-enhancing methods. For example, in the scenario in Figure \ref{fig:data-partition}, only the financial institution has access to the sensitive attributes values. Secondly, organizations involved in real-world VFL systems often have varying computational capabilities and may complete their local updates at different speeds. Requiring each organization to perform a single local update per communication round when training a fair model can lead to inefficiencies \cite{liu2022achieving}.

Despite its importance, research specifically addressing group fairness in VFL remains limited, with only one work presented. Liu \textit{et al.} \cite{liu2022achieving} examine a VFL scenario where data parties and a central server collaborate to train a machine learning model, with each feature vector distributed across the data parties. They identify two types of data parties: active parties, which initiate the task and possess information about labels, sensitive attributes, and the loss function, and passive parties, which do not have access to this information. The server is assumed to have access to both the labels and sensitive attributes. To address the challenge of imbalanced computational resources, they allow each active data party to perform multiple local gradient updates in parallel before exchanging information with the server. For passive parties, a single model update is conducted between two consecutive communication rounds with the server.

\subsection{Federated Transfer Learning}\label{sec:transfer}

Federated Transfer Learning (FTL) is an extension of FL that leverages knowledge from a source domain to improve the learning process in a target domain where data might be scarce or unlabeled \cite{flsurvey, saha2021federated}. In FTL, the participating clients in the source domain have abundant labeled data, while those in the target domain may have limited or no labeled data. The goal is to transfer the knowledge gained from the source domain to the target domain.

FTL is particularly useful in scenarios where direct FL might not be feasible due to the lack of adequate data in the target domain. For instance, consider a financial institution, \textit{A}, that wants to develop a mortgage approval model. Some institution, \textit{B}, may have extensive data on general loan applications and customer credit histories (source domain), while \textit{A} might only have limited data on specific to mortgage approvals (target domain). FTL can facilitate the transfer of knowledge from the well-established credit scoring models based on general loan data to enhance the performance of models tailored for mortgage approvals, for example. This way, institutions with limited data can benefit from models trained on broader datasets.

Achieving group fairness in FTL presents several unique challenges. Firstly, the source and target domains may have different distributions of sensitive attributes. Ensuring fairness across these domains is challenging, as the transferred knowledge might introduce or exacerbate biases in the target domain. Moreover, defining and measuring fairness in the context of FTL is complex, as the metrics used in the source domain may not be suitable for the target domain.

Despite the importance of these challenges, no dedicated work has been proposed to address group fairness in FTL specifically. This gap highlights a significant opportunity for future research to develop novel strategies that ensure fairness in FTL while maintaining the privacy of the learning process.

\section{Location}\label{section:location}

\begin{figure*}[h!]
    \centering
    \includegraphics[width= \linewidth]{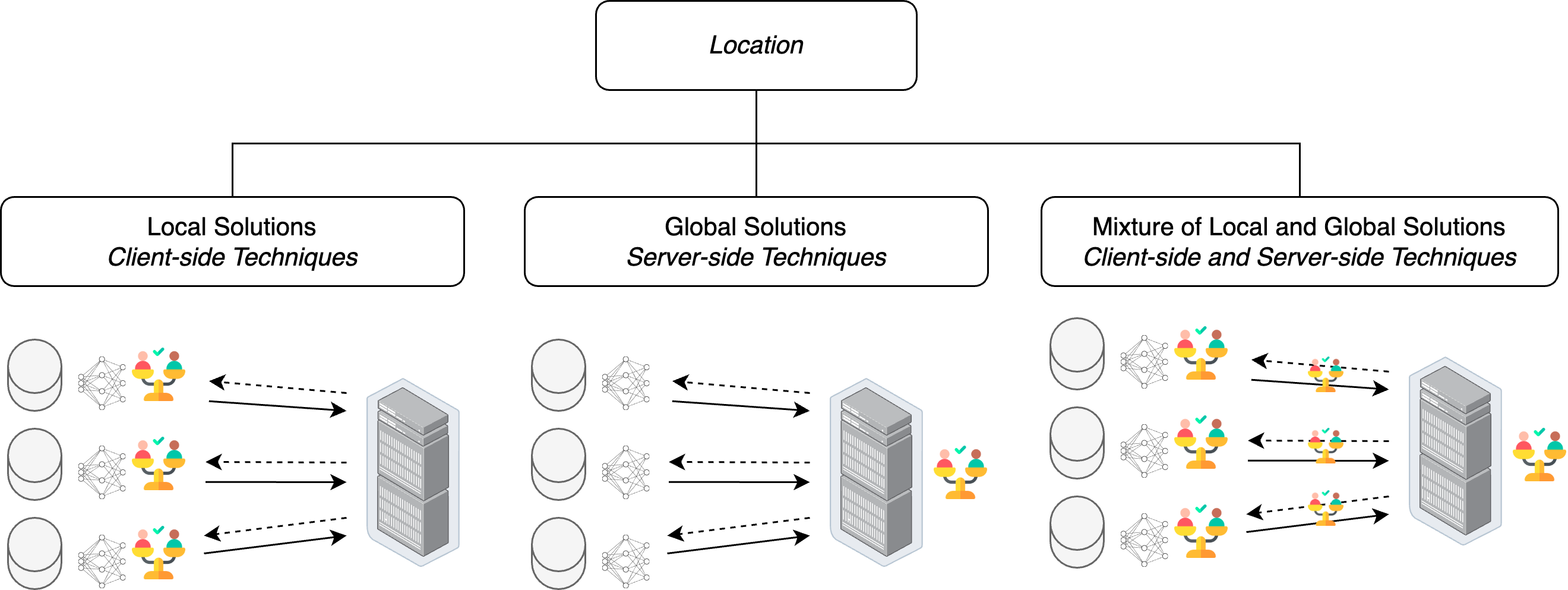}
    \caption{Types of fair Federated Learning approaches categorized by location.}
    \label{fig:location}
\end{figure*}

In this section, we introduce a novel categorization of current approaches to achieving group fairness in FL into three main types based on where the fairness operations are conducted: local methods, global methods, and a mixture of local and global methods. Each approach has its advantages and disadvantages, which are discussed in detail in the following subsections. Figure \ref{fig:location} presents the types of fair FL approaches categorized by location.

\subsection{Local Solutions / Client-side Techniques}

Local solutions focus on implementing fairness-aware strategies on each client independently, without direct intervention from the central server. These methods leverage the clients' local data to mitigate biases and ensure fairness locally.

One advantage of local methods is the preservation of data privacy. Since all fairness adjustments are made locally, there is no need to share sensitive data with the central server. In addition, these methods allow clients to tailor fairness strategies specifically to their own local data, which can potentially align fairness objectives with the unique context of each client.

However, local methods face several challenges. They often struggle to achieve global fairness, especially when clients' data distributions are non-IID and do not adequately represent the global population \cite{ezzeldin2023fairfed}. This limitation can lead to a lack of cohesion in achieving fairness across the entire model. Additionally, local fairness objectives may conflict with each other, creating difficulties in reconciling these discrepancies to ensure global fairness. Furthermore, some clients may only have data from specific groups rather than all groups, making it challenging to achieve fairness across the global model. This can result in unequal representation and potential biases in the global model.

These challenges emphasize the need for approaches that integrate the server to achieve a more comprehensive solution for group fairness in FL.

\subsection{Global Solutions / Server-side Techniques}

Global solutions rely on the central server to implement fairness-aware strategies, aggregating model updates from clients in a way that promotes fairness. These methods do not require direct involvement of the clients in the fairness process.

Similarly to local methods, the main advantage of these methods is privacy, as no sensitive client information needs to be shared with the central server beyond the necessary model updates. Furthermore, one advantage of global methods is their independence from any client-side procedures, meaning that clients do not need to implement any fairness strategies locally. This simplicity reduces the burden on clients, as the central server takes care of all fairness-related adjustments. Additionally, the server can use model information and a validation set to evaluate fairness and weight clients accordingly, leading to a more globally fair model.

However, these methods also have drawbacks. They might overlook local data nuances specific to individual clients, making it more difficult to ensure fairness at the local level. This can lead to less effective fairness adjustments for certain data distributions. Moreover, in this setting the central server becomes a single point of failure and may face scalability issues as the number of clients increases.

\subsection{Hybrid Solutions / Client-side and Server-side Techniques}

The majority of current solutions involve a hybrid solution, where clients and the central server collaborate to achieve fairness. These hybrid approaches aim to leverage the strengths of both local and global strategies while mitigating their respective weaknesses.

Despite their effectiveness, these methods introduce additional privacy concerns, as clients may need to share extra information with the server, such as intermediate fairness-related statistics, demographic information, or performance metrics. Sharing such sensitive information could potentially compromise client privacy, which is an important concern in FL. To mitigate these risks, hybrid methods often incorporate privacy-preserving mechanisms, such as differential privacy or secure multi-party computation. These techniques ensure that the extra information shared by clients remains private while still allowing the server to make fairness-aware updates to the global model.

\section{Strategies}\label{sec:strategies}

In this section, we explore the various strategies used to achieve group fairness in FL. Table \ref{tab:strategies} provides an overview of these strategies along with their respective works.

\begin{table}
\centering
\begin{tabular}{p{2.5cm}|p{6.5cm}|p{3cm}}
\toprule
\textbf{Strategy} & \textbf{Description} & \textbf{Works} \\
\toprule
Aggregation & adjusting the weight of each client's update & \cite{kanaparthy2022fair} \cite{mehrabi2022towards} \cite{salazar2023fair} \cite{djebrouni2024bias} \\
\midrule
Reweighting & adjusting the influence of specific datapoints during training & \cite{abay2020mitigating} \cite{du2021fairness} \cite{du2021fairness} \cite{zeng2022improving} \cite{zeng2022improving} \cite{ezzeldin2023fairfed} \\
\midrule
\parbox[t]{3cm}{Adversarial \\ Learning} & training models in a competitive environment & \cite{liang2020think} \cite{yang2022towards} \cite{han2024fair} \\
\midrule
\parbox[t]{3cm}{Client \\ Participation} & selecting or rewarding clients based on their contributions to the model & \cite{zhang2020fairfl} \cite{salazar2023fair} \\
\midrule
Personalization & adapting the global model to better meet the needs of individual clients & \cite{carey2022robust} \cite{agrawal2024no} \cite{wang2024analyzing} \\
\midrule
Clustering & grouping clients based on shared characteristics & \cite{kyllo2023inflorescence} \cite{yang2023enhancing} \cite{nafea2022proportional} \cite{meerza2024glocalfair} \cite{salazar2024unveiling}  \\
\midrule
Thresholding & adjusting decision boundaries or modifying prediction outputs & \cite{pentyala2022privfairfl} \cite{yin2024distribution} \\
\midrule
\parbox[t]{3cm}{Constrained \\ Optimization} & optimizing an objective function subject to a set of constraints & \cite{chu2021fedfair} \cite{rodriguez2021enforcing} \cite{padala2021federated} \cite{agrawal2024no} \cite{cui2021addressing} \cite{papadaki2022minimax} \cite{hu2022fair} \cite{selialia2023mitigating} \cite{badar2024fairtrade} \cite{su2024multi} \cite{liu2022achieving} \\
\bottomrule
\end{tabular}
\caption{Works on group fairness in Federated Learning categorized by strategy.}
\label{tab:strategies}
\end{table}

\subsection{Aggregation}\label{sec:aggregation}

Aggregation strategies, which are global solutions, do not need the direct involvement of clients in ensuring fairness \cite{salazar2023fair, mehrabi2022towards}. By adjusting the weight of each client's update based on fairness considerations, aggregation strategies can mitigate biases that might arise from uneven data distributions or varying levels of client participation.

Aggregation strategies typically use a validation or proxy dataset, which allows the server to compute fairness measures concerning existing sensitive attributes in the data. As explained in Section \ref{section:identifying-benchmarking}, clients can contribute to creating this validation set, ensuring that it is representative of the overall data distribution.

Kanaparthy \textit{et al.} \cite{kanaparthy2022fair} propose simple strategies for aggregating local models based on different heuristics, such as using the model with the least fairness loss for a given fairness notion or the model with the highest accuracy-to-fairness loss ratio on the validation set. Similarly, Mehrabi \textit{et al.} \cite{mehrabi2022towards} introduce FedVal, a simple aggregation strategy that performs a weighted average of client models based on fairness or performance measures on a validation set. This approach accommodates multiple fairness metrics with weights for trade-offs, although it faces challenges with uncooperative clients. 

Distinct from these, T. Salazar \textit{et al.} \cite{salazar2023fair} propose FAIR-FATE, which uniquely combines two types of updates: one focused on performance and the other on fairness, using a decaying momentum \cite{fernandes2021decay} strategy to balance these over time. This approach aims to achieve fairness by gradually shifting towards fairer updates while addressing fluctuations in gradients that are biased.

Finally, Djebrouni \textit{et al.} \cite{djebrouni2024bias} propose Astral, which stands out from other methods by using an evolutionary algorithm to guide the aggregation process. This approach leverages a proxy dataset to reweight clients based on their contributions to fairness, ensuring that model bias remains below an adjustable threshold while continuously maximizing accuracy.

\subsection{Reweighting}

Reweighting is a machine learning strategy that adjusts the influence of datapoints during training by assigning different weights to them based on specific criteria. In the context of fairness, to ensure equitable outcomes these criteria may include group distributions and counts or the result of fairness metrics' evaluations.

The first work to propose a reweighting strategy was by Abay \textit{et al.} \cite{abay2020mitigating}. They introduce a local reweighting solution based on the method from \cite{kamiran2012data}, where each client calculates weights as the ratio of the expected probability to the observed probability of the sample's sensitive attribute. These weights are applied locally by each client, allowing them to avoid sharing sensitive attributes or data sample information with the aggregator. However, this method may not achieve global fairness objectives and can be less effective if the clients' sensitive attribute distributions are non-IID, making it less practical in real-world scenarios \cite{ezzeldin2023fairfed}. Additionally, Abay \textit{et al.} \cite{abay2020mitigating} propose a hybrid solution - global reweighting - which uses the combined information from all clients. In this approach, if clients agree to share their sensitive attributes' sample counts with the aggregator, a differentially private global reweighting method can be employed. The server collects the statistics with differential privacy noise, computes the global reweighting weights, and then distributes these weights back to the clients, who use them during training.

Du \textit{et al.} \cite{du2021fairness} introduce AgnosticFair, which uniquely addresses the challenge of handling unknown test data distributions. Different from other methods, AgnosticFair uses kernel reweighing functions to assign a reweighing value on each training sample in both loss function and a fairness constraint. To ensure robustness against varying data distributions, they frame the problem as a two-player adversarial minimax game between a learner and an adversary. In this setup, the adversary generates potential unknown test data distributions to maximize the classifier's loss, whereas the learner tries to find parameters to minimize the worst case loss over the unknown testing data distribution produced by the adversary. 

Instead of assigning individual reweighting values to each training sample \cite{du2021fairness}, Zeng \textit{et al.} \cite{zeng2022improving} adjust the weight of the local loss function for each sensitive group. They propose FedFB to extend FairBatch \cite{fairbatch} to an FL setting. Specifically, they modify the FedAvg algorithm so that each client shares not only its models but also its fairness statistics with the server. Once the server receives the securely aggregated model parameters and fairness statistics, it performs both model averaging and updating of reweighting coefficients. The server then broadcasts the averaged model parameters together with the updated coefficients, which are then used for the subsequent round of local training with a reweighted loss function. 

Similar to FedFB \cite{zeng2022improving}, Ezzeldin \textit{et al.} \cite{ezzeldin2023fairfed} employ FedAvg and a reweighting mechanism. They propose FairFed \cite{ezzeldin2023fairfed}, where the client coefficients are adaptively adjusted based on the deviation of each client's fairness metric from the global average. The clients evaluate the fairness of the global model on their local datasets in each round and collectively collaborate with the server to adjust its model weights. The weights are a function of the mismatch between the global fairness measurement (on the full dataset) and the local fairness measurement at each client, favoring clients whose local measures match the global measure.

\subsection{Adversarial Learning}

Adversarial learning is a machine learning technique that involves training models in a competitive environment, where one model (often referred to as the generator or adversary) attempts to deceive another model (discriminator or classifier). In this setup, the adversary is designed to create challenging scenarios or examples that are difficult for the classifier to handle. The classifier, in turn, learns to improve its performance by correctly classifying these adversarial examples. 

In the context of fairness, adversarial learning can be employed to train an adversary to make machine learning models invariant to sensitive attributes or a generator to create novel examples that enhance the diversity of the training data. Although adversarial learning is widely used in centralized learning to mitigate bias, extending it to a federated framework presents significant privacy and convergence challenges.

The first work to apply adversarial learning to achieve group fairness in FL was by Liang and Liu \textit{et al.} \cite{liang2020think}. They proposed Local Global Federated Averaging (LG-FEDAVG), a method that simultaneously learns high level local representations on each device while training a global model across all devices. To ensure fair local representations, they used adversarial training, enabling the local models to produce distributions that are invariant with respect to these attributes.

Different from this approach, Yang \textit{et al.} \cite{yang2022towards} present FairSCAT, a semi-centralized adversarial training approach that uses a Variational AutoEncoder (VAE) tailored for FL environments. In their approach, the VAE decoder is maintained on the server side, while the encoder remains on the client side. The server sends the encoder parameters of a pretrained VAE model to the clients. Clients generate adversarial samples locally, compress them into latent variables, and manipulate part of these based on the sensitive attribute to create adversarial feature dimensions. In each training round, these feature dimensions are sent to the server, which uses the VAE decoder to reconstruct adversarial samples. The server then trains a federated model using these adversarial samples to improve group fairness.

Finally, Han \textit{et al.} \cite{han2024fair} introduce FFL-OppoGAN, a novel method leveraging Generative Adversarial Networks (GANs) to produce synthetic tabular data with features that are opposite to those in the original dataset. This approach incorporates a fairness constraint directly into the generator's loss function, ensuring that the generated data promotes equitable outcomes. To enhance the quality of the generated data and address common challenges in GANs, such as mode collapse, OppoGAN uses a Wasserstein GAN. Additionally, the method also ensures performance distribution fairness.

\subsection{Client Participation}

Client participation techniques in FL involve strategically selecting or rewarding clients based on their contributions to the model. For group fairness, these techniques can prioritize clients whose data and updates enhance fairness across sensitive groups, ensuring that the federated model does not disproportionately favor or disadvantage any particular group.

Zhang \textit{et al.} \cite{zhang2020fairfl} propose FairFL, a deep multi-agent reinforcement learning framework that optimizes both fairness and accuracy in FL. They introduce novel reward and state functions that guide clients in collaboratively making local update decisions that enhance the global model's fairness. Their approach trains a client-selection policy function using multi-agent reinforcement learning, maximizing a gain function focused on bias mitigation in the global model.

Another approach, FAIR-FATE \cite{salazar2023fair} by T. Salazar et. al, uses a validation set to evaluate and select specific clients for model updates. This technique prioritizes contributions from clients whose local updates demonstrate higher fairness compared to the current global model, ensuring that only the most equitable updates are incorporated into the federated model.

\subsection{Personalization}

Personalization in FL refers to tailoring the global model to better meet the needs of individual clients by adapting it to their specific data distributions and requirements. In the context of group fairness, personalization can improve fairness by adjusting models to account for variations in data related to sensitive attributes, as well as the diverse demands of different clients. Additionally, it can mitigate biases that may arise from a one-size-fits-all approach. In this setup, to evaluate group fairness it is important to compare the performance of the personalized models with the global model. If certain groups experience a significant drop in performance after personalization, it may indicate bias in the model's ability to serve those groups effectively. By comparing fairness metrics before and after personalization, the shift in fairness can help identifying whether personalization improves or reduces fairness.

Carey \textit{et al.} \cite{carey2022robust} investigate fairness heterogeneity, which arises when clients enforce different fairness metrics during local training. They introduce Fair Hypernetworks (FHN), a personalized FL framework that accommodates varying fairness requirements and performs robustly in non-IID settings. FHN allows each client to select its own fairness metric by incorporating these metrics as linear constraints in the local optimization function. Hypernetworks, which generate network parameters for other models, are well-suited for this task since they can produce a range of personalized models tailored to each client's specific fairness criteria. This flexibility makes FHN effective in managing conflicting fairness constraints across clients. 

Agrawal \textit{et al.} \cite{agrawal2024no} propose F\textsuperscript{2}PGNN (Fair Federated Personalized Graph Neural Network), a framework that integrates personalized Graph Neural Networks (GNNs) with fairness considerations. In this approach, each client maintains its user-item interaction history to construct a local subgraph. Clients then train a GNN by incorporating higher-order information into their respective subgraphs. To ensure fairness, a fairness constraint is added as a regularizer in the objective function, meaning fairness is explicitly enforced rather than emerging solely from personalization, although personalization can enhance it.

Regarding specific group fairness applications, Wang \textit{et al.} \cite{wang2024analyzing} assess the impact of personalized FL on group fairness within the healthcare domain, using two real-world Electronic Health Record (EHR) datasets. Their findings show that, on average, personalized FL models achieve better fairness compared to standalone training. Additionally, while personalized models and the global model provide comparable fairness benefits for most hospitals, these benefits vary across institutions. Their work shows that personalization tends to improve fairness in hospitals with more significant bias issues but can exacerbate fairness problems in hospitals with less biased data.

\subsection{Clustering}

In FL, clustering clients based on shared characteristics can enhance the model's fairness by tailoring updates to specific groups. This technique is particularly valuable for improving group fairness, as it allows for the management of client diversity and the addressing of fairness-related concerns. While clustering is often considered a form of personalization, as it tailors models to groups of clients, we distinguish it from true personalization. Personalization typically refers to adapting the model for each individual client, whereas clustering involves grouping multiple clients based on different characteristics.

Several works have focused on understanding the fairness implications of clustering without explicitly designing for fairness \cite{kyllo2023inflorescence, yang2023enhancing}. In particular, Kyllo \textit{et al.} \cite{kyllo2023inflorescence} analyze how clustered FL strategies that do not incorporate fairness mechanisms affect fairness outcomes. They found that while these methods improve certain fairness metrics, such as accuracy equality, they are less effective at addressing more challenging fairness criteria such as disparate impact and equalized odds. This highlights the limitations of fairness-unaware clustering and suggests the need for more fairness-focused clustering approaches.

Other works have explicitly developed clustering strategies with fairness objectives. Nafea \textit{et al.} \cite{nafea2022proportional} introduce IFFCA that ensures proportional representation of protected groups in each cluster. This method integrates fairness directly into the clustering process by making cluster assignments based on proportional fairness and using unsupervised techniques to balance learning performance with fairness. IFFCA stands out for its focus on proportionality, ensuring that minority or protected groups are adequately represented in the learning process.

Meerza \textit{et al.} \cite{meerza2024glocalfair} also introduce a fairness-centric clustering approach with GLocalFair, which clusters clients based on their fairness levels as measured by the Gini Coefficient, which serves as a fairness proxy. To update the global model, they calculate a weighted mean within each cluster based on their dataset size.

Finally, T. Salazar \textit{et al.} \cite{salazar2024unveiling} propose the FairFedDrift algorithm to deal with group-specific distributed concept drift, clustering clients together over time based on shared concepts. Their approach emphasizes the limitations of single global models when dealing with distributed drifts.

\subsection{Thresholding}

Thresholding is a post-processing technique commonly used in fairness in machine learning to adjust decision boundaries or modify prediction outputs to meet fairness criteria. This method involves setting thresholds on model outputs to ensure that certain fairness metrics are met. In the context of FL, thresholding can similarly be used to improve group fairness.

Pentyala \textit{et al.} \cite{pentyala2022privfairfl} propose PrivFairFL-Post, a privacy-preserving technique that identifies fair classification thresholds for different groups in FL. PrivFairFL-Post is applied after the training phase, assuming that each client has already received the final model. After training, clients generate prediction probabilities and share encrypted sensitive data with secure multi-party computation servers. The servers then construct noisy ROC curves for protected and unprotected groups and the optimal thresholds are computed and shared with clients.

Similarly, Yin \textit{et al.} \cite{yin2024distribution} propose FedFaiREE, a post-processing technique that also relies on the concept that achieving optimal misclassification performance under specific fairness constraints requires setting different thresholds for different groups. However, different from PrivFairFL-Post \cite{pentyala2022privfairfl}, FedFaiREE uses distributed order statistics to enforce these fairness constraints and selects the classifier with the highest accuracy among those that meet the criteria.

\subsection{Constrained Optimization}

Constrained optimization involves optimizing an objective function subject to a set of constraints. In the context of group fairness in FL, constrained optimization is used to ensure that models not only perform well on average but also adhere to fairness criteria across different client groups. 

One approach to achieving global fairness in FL is to formulate a constrained optimization problem where each client seeks to optimize their local model while ensuring that fairness-related disparities do not exceed a predefined threshold. The models are then aggregated to form a global model that balances accuracy and fairness across all clients \cite{chu2021fedfair, rodriguez2021enforcing, padala2021federated, agrawal2024no}. Alternatively, some methods employ bi-level optimization, aiming to identify the global model with the lowest overall loss while minimizing the worst-case fairness violation across clients \cite{cui2021addressing, papadaki2022minimax, hu2022fair, selialia2023mitigating}.

Different strategies are employed to solve these constrained optimization problems. For instance, Chu \textit{et al.} \cite{chu2021fedfair} propose a method that introduces a fairness constraint using a Lagrangian multiplier, converting the problem into a nonconvex-concave min-max problem addressed by the Alternating Gradient Projection algorithm. Rodríguez-Gálvez \textit{et al.} \cite{rodriguez2021enforcing} adapt the method of differential multipliers with a quadratic penalty term to enforce fairness. Padala \textit{et al.} \cite{padala2021federated} use a two-phase approach: first, they apply fairness constraints as a regularization term in the loss function; then, they train a surrogate model that replicates the fair predictions while ensuring differential privacy.

Other methods focus on optimizing fairness through more complex formulations. Cui \textit{et al.} \cite{cui2021addressing} frame the problem as constrained multi-objective optimization, achieving Pareto optimality by controlling the gradient direction. Papadaki \textit{et al.} \cite{papadaki2022minimax} propose a minimax approach that weighs empirical loss by a trainable vector and finds the optimal model for the worst-case scenario, making it suitable for cases where clients have access to only a subset of population groups. Hu \textit{et al.} \cite{hu2022fair} extend this by introducing a bounded group loss constraint, where the loss for each group is capped, claiming that the Papadaki method is a special case of their approach when certain hyperparameters are fixed. Selialia \textit{et al.} \cite{selialia2023mitigating} compute group importance weights to scale losses and introduce regularized multiplicative weight updates to mitigate bias, along with methods to set performance thresholds for different groups. Badar \textit{et al.} \cite{badar2024fairtrade} present a multi-objective optimization framework that balances balanced accuracy and fairness using Differentiable Expected q-Hypervolume Improvement. Su \textit{et al.} \cite{su2024multi} address fairness by detecting and adjusting gradient conflicts across clients before aggregation, ensuring that conflicting gradients do not compromise fairness.

In contrast to these HFL approaches, Liu \textit{et al.} \cite{liu2022achieving} tackle fairness in VFL using an asynchronous gradient coordinate-descent ascent algorithm.

\section{Concerns}\label{section:concerns}

Achieving group fairness in FL is a complex challenge, with several concerns that researchers and practitioners must navigate. These concerns stem from the unique characteristics of FL, such as its decentralized nature and the diversity of data across clients. In this section, we introduce several concerns that are important to addressing group fairness in FL, including issues related to non-IID data, privacy, robustness, and concept drift.

\subsection{Non-IID}

Due of its decentralized nature, FL exacerbates the problem of bias since clients' data distributions can be very heterogeneous. This heterogeneity means that some clients may have a high representation of datapoints from specific sensitive groups, while others may have very low or no representation of those groups. Consequently, it is common in FL research to study algorithms under different non-IID settings \cite{flsurvey}.

In the context of group fairness, Ezzeldin \textit{et al.} \cite{ezzeldin2023fairfed} were the first to investigate this issue using a non-IID synthesis method based on the Dirichlet distribution, which allows for configurable sensitive attribute distributions. Building on this, T. Salazar \textit{et al.} \cite{salazar2023fair} also applied this method, incorporating both sensitive and target distributions. Specifically, for each sensitive attribute value $s$ and target value $y$, they sample $p_{s,y} \sim Dir(\sigma)$ and allocate a portion $p_{s,y,k}$ of the datapoints with $S = s$ and $Y = y$ to client $k$. The parameter $\sigma$ controls the heterogeneity of the distributions in each client, where $\sigma \rightarrow \infty$ results in IID distributions. Additionally, some studies, such as Papadaki \textit{et al.} \cite{papadaki2022minimax}, further explore scenarios where certain clients have no representation of a particular group.

\begin{figure}[h]
    \centering
    \subfloat[\centering $\sigma=0.5$]{{\includegraphics[width=5.75cm]{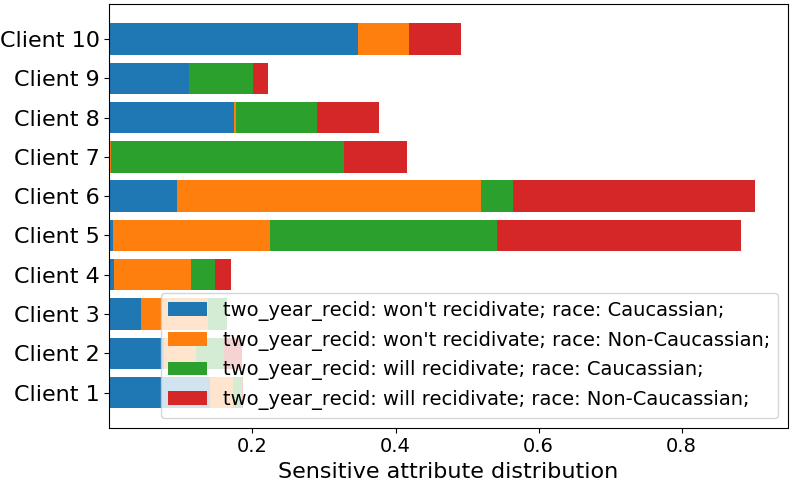}}}%
    \qquad
    \subfloat[\centering $\sigma \rightarrow{} \infty$]{{\includegraphics[width=5.75cm]{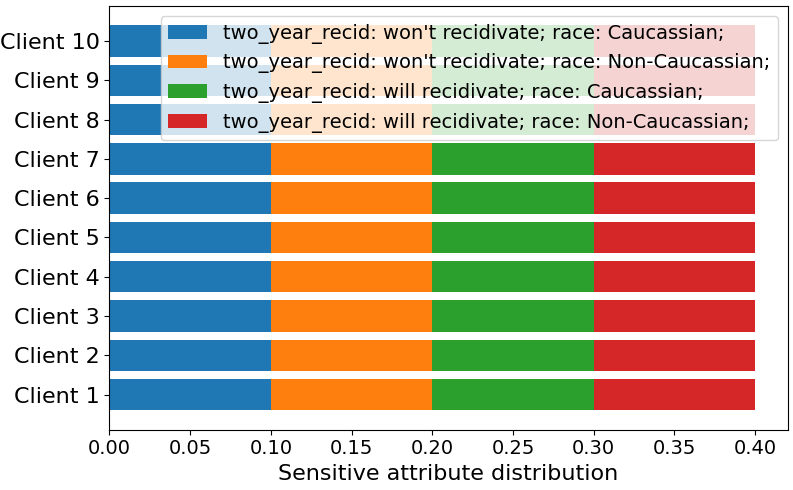}}}%
    \caption{Examples of heterogeneous data distributions on the COMPAS \cite{compas} dataset using race as the sensitive feature for 10 clients with $\sigma=0.5$ and $\sigma \rightarrow{} \infty$ \cite{salazar2023fair}}%
    \label{fig:distributions}%
\end{figure}

To illustrate, Figure \ref{fig:distributions} \cite{salazar2023fair} presents examples of heterogeneous data distributions on the COMPAS \cite{compas} dataset using race as the sensitive feature for 10 clients with $\sigma=0.5$ and $\sigma \rightarrow \infty$. For $\sigma=0.5$, it can be observed that different clients have very different representations of protected and unprotected groups, as well as positive and negative outcomes. For example, while client 10 has about 40\% of Caucasians who won't recidivate, client 9 has only about 10\%. Furthermore, different clients have varying numbers of datapoints. In contrast, for $\sigma \rightarrow \infty$, the representations of different groups across clients are uniform.

Overall, studying the impact of non-IID settings is crucial for developing fair FL algorithms, as it reflects the real-world scenarios where data distributions across clients are rarely homogeneous.

\subsection{Privacy}

Privacy is a fundamental concern in FL. While FL inherently provides some level of data privacy by keeping data localized on clients' devices, the incorporation of group fairness methods can involve sharing more than just model updates, potentially increasing the risk of privacy breaches. Thus, it is important to take extra precautions to preserve privacy when implementing fairness techniques in FL. To address these privacy concerns, several works have introduced different mechanisms to enhance the privacy of their approaches.

Differential Privacy (DP) \cite{dwork2006differential} is a popular technique for ensuring that the inclusion or exclusion of a single datapoint does not significantly affect the outcome of any analysis, thus preserving the privacy of individual datapoints. Abay \textit{et al.} \cite{abay2020mitigating}, Gu \textit{et al.} \cite{gu2022privacy}, Sun \textit{et al.} \cite{sun2023toward}, and Agrawal \textit{et al.} \cite{agrawal2024no} apply DP to enhance the privacy of FL models while ensuring fairness and protecting sensitive information. In addition to the above methods, Chen \textit{et al.} \cite{chen4601101enforcing} focus on localized forms of DP to enhance privacy in fair FL.

Secure Multiparty Computation (SMC) \cite{truex2019hybrid} is another technique used to enhance privacy in FL. SMC allows multiple parties to collaboratively compute a function over their inputs while keeping those inputs private. Zhang \textit{et al.} \cite{zhang2020fairfl} leverage SMC to ensure that the computations involved in FL are performed securely, preserving the privacy of each client's data. Furthermore, Padala \textit{et al.} \cite{padala2021federated}, Ezzeldin \textit{et al.} \cite{ezzeldin2023fairfed}, and Pentyala \textit{et al.} \cite{pentyala2022privfairfl} combine DP and SMC to achieve stronger privacy guarantees.

Overall, while FL offers inherent privacy benefits, the integration of group fairness methods requires additional privacy considerations. By employing techniques such as DP, SMC, or a combination of both, researchers can enhance the privacy of FL models, ensuring that fairness does not come at the cost of data confidentiality.

\subsection{Robustness}\label{sec:robustness}

Robustness refers to a system's ability to withstand various types of adversarial attacks while maintaining reliable performance \cite{lyu2022privacy}. Robustness in FL is important to ensure that the aggregated model is not influenced by malicious or faulty clients. A particularly challenging aspect of robustness in FL is safeguarding against poisoning attacks, where malicious clients intentionally submit incorrect updates to degrade model performance or skew it towards biased outcomes. This challenge becomes even more complex when considering group fairness, as it is essential to distinguish between honest minority group members and potential model poisoners.

Touat \textit{et al.} \cite{touat2023towards} explore the intersection of robustness and fairness in FL. They show that classical robust FL methods may inadvertently filter out benign clients with statistically rare data, particularly affecting minority groups. Traditional robust methods often misinterpret updates from minority groups as anomalies or potential attacks, leading to the unfair exclusion of these clients from the aggregation process.

Developing robust FL mechanisms that account for the statistical rarity and heterogeneity of data from minority groups is important for enhancing both the fairness and robustness of federated models.

\subsection{Concept Drift}\label{sec:concept}

Ensuring group fairness in FL under concept drift presents several open challenges. Concept drift refers to the scenario when the relation between the input data and the target variable changes over time \cite{gama2014survey}. This phenomenon can significantly impact the performance and fairness of machine learning models, as they may become less accurate and more biased as the underlying data distribution changes.

T. Salazar \textit{et al.} \cite{salazar2024unveiling} introduce the problem of group-specific distributed concept drift in FL. Group-specific distributed concept drift occurs when different clients in a FL setting experience distinct group-specific concept drifts. Specifically, group-specific concept drift refers to the situation where one group's conditional distribution of the target variable changes over time, while other groups' conditional distributions remain constant. These temporal and spatial dynamics can lead to significant challenges in maintaining both fairness and accuracy over time. T. Salazar \textit{et al.} \cite{salazar2024unveiling} propose the FairFedDrift algorithm to address this issue which uses a multi-model approach, a local group-specific drift detection mechanism, and continuous clustering of models over time.

Ensuring fairness in the presence of concept drift, particularly when it affects different groups unequally, is important for maintaining the fairness of FL systems. Addressing group-specific drifts requires continuous monitoring to account for evolving patterns in the data. Future work in this area could explore more adaptive algorithms and real-time drift detection mechanisms that ensure fairness in dynamic FL environments.

\section{Sensitive Attributes}\label{section:sensitive-attributes}

\begin{figure}[h!]
    \centering
    \includegraphics[width= \linewidth]{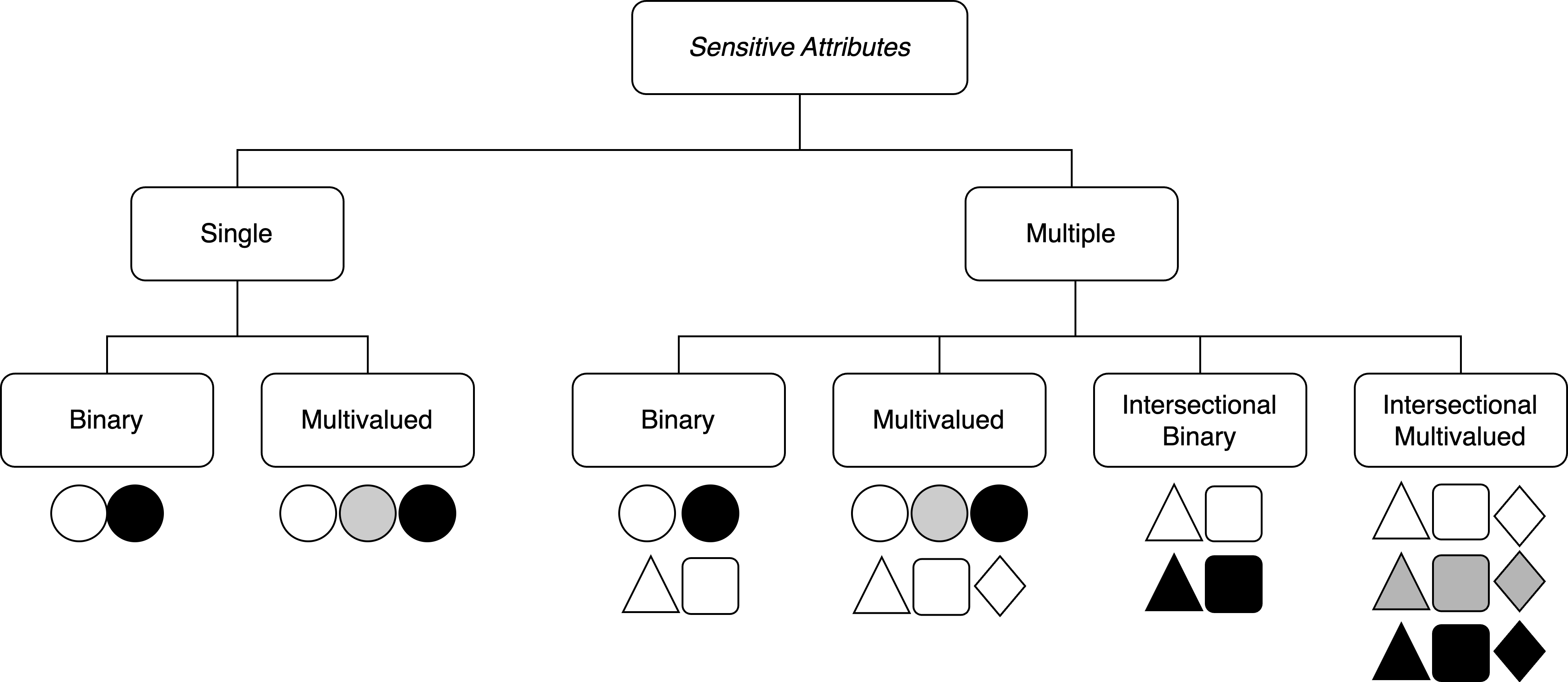}
    \caption{Different ways to consider sensitive attributes when accounting for group fairness.}
    \label{fig:sensitive-attributes}
\end{figure}

In the area of fair machine learning, sensitive attributes are factors that need careful consideration to ensure equitable outcomes. Figure \ref{fig:sensitive-attributes} presents different ways to consider groups within sensitive attributes. 

Sensitive attributes can be evaluated either individually or in combination with others. When focusing on a single attribute, one can consider its binary representation (e.g., male/female) or its multi-valued form (e.g., various racial categories). Furthermore, when examining multiple sensitive attributes simultaneously, it becomes possible to explore their intersections.

Typically, many works on group fairness focus on binary versions of sensitive attributes. This binary approach can be problematic since it may oversimplify the diversity of groups. It is important to consider multiple groups within the same axis (multivalued sensitive attributes) to capture a more nuanced understanding of fairness issues. In addition, Wang \textit{et al.} \cite{wang2022towards} point out that within a single axis, groups that are quite diverse may come together not because they share a specific characteristic, but because they face similar challenges and have joined forces to advocate for change. For example, the category `Disability' includes individuals with a wide range of different impairments. These individuals may not share the same type of disability, but they unite since they face similar societal obstacles and discrimination, motivating them to work together for greater inclusion and equity.

Furthermore, most research in fair machine learning focuses on single sensitive attributes, or, when considering multiple attributes, it treats them independently rather than examining their combined impact by considering intersectionality. The term `intersectionality', defined by Crenshaw \cite{crenshaw1989demarginalizing}, highlights how different identities along various axes intersect to produce unique forms of discrimination and societal effects. For example, Black women may experience discrimination that is not merely the sum of being Black and being female, but a unique combination of both identities.

Handling multiple sensitive attributes and their intersections presents significant challenges. The more identities we consider, the smaller each subgroup becomes, which can introduce computational burdens during model training. Despite these challenges, understanding the structure within intersectional data can be advantageous. For instance, in certain circumstances, learning about statistical patterns in an under-represented Black Female group from groups it might share characteristics with, such as Black Males, can be beneficial \cite{wang2022towards}.

This problem is further intensified in FL since data is distributed across multiple clients, each with potentially different distributions of sensitive attributes. The majority of works in FL only consider a single binary sensitive attribute. Looking at Table \ref{tab:strategies} it is possible to identify that only three works \cite{liang2020think, meerza2024glocalfair, djebrouni2024bias} have explored fairness with multiple attributes. Furthermore, only nine works have considered multivalued sensitive attributes \cite{zhang2020fairfl, rodriguez2021enforcing, kanaparthy2022fair, zhang2022unified, papadaki2022minimax, zeng2022improving, poulain2023improving, wang2024analyzing, su2024multi}. Despite these efforts, no research work has addressed intersectionality in the context of FL.

To sum up, the field still lacks comprehensive approaches to handle intersectionality in FL. Addressing this gap is crucial for developing fair FL models that account for the complex, intersecting identities of individuals.

\section{Datasets and Applications}\label{section:datasets-and-applications}

Table \ref{tab:datasets} provides a comprehensive summary of the datasets used in fair FL research. It analyses them based on their data type, application areas, target outcomes, size, and the studies from Table \ref{tab:strategies} that use each dataset. By examining these datasets, we can gain insights into the current landscape of fairness research and identify opportunities for expanding into underexplored data types and applications.

Many works on group fairness in FL use traditional fair machine learning datasets such as Adult \cite{adult}, KDD \cite{kdd}, Credit Card Default \cite{credit-card-default}, Dutch \cite{dutch}, Bank Marketing \cite{bank}, COMPAS \cite{compas}, Law \cite{law}, and Crime \cite{crime}. These datasets contain sensitive groups such as gender, age or race. For a detailed explanation of these datasets, we refer to \cite{datasets-fair-ml}. 

Some of these datasets have been overused. In particular, the Adult dataset \cite{adult} which is derived from a 1994 US Census survey, has been used in 33 out of the 48 works in fair FL. However, researchers have identified several idiosyncrasies that limit its external validity \cite{acs-datasets}, such as its age and outdated feature encodings. To address these limitations, a suite of new datasets \cite{acs-datasets} derived from US Census surveys has been introduced, providing prediction tasks related to income, employment, health, transportation, and housing.

\begin{small} 
\begin{longtable}{p{1.6cm}|p{0.75cm}|p{1.9cm}|p{1.7cm}|p{1.3cm}|p{1.8cm}|p{1.4cm}}
\caption{
Summary of datasets used in works on group fairness in Federated Learning.
}
\label{tab:datasets} \\
\toprule
\textbf{Dataset} & \textbf{Type} & \textbf{Area} & \textbf{Target} & \textbf{Size} & \textbf{Works} & \textbf{\% Usage in 48 works} \\
\toprule
\endfirsthead
\toprule
\textbf{Dataset} & \textbf{Type} & \textbf{Area} & \textbf{Target} & \textbf{Size} & \textbf{Works} & \textbf{\% Usage in 48 works} \\
\toprule
\endhead
Adult \cite{adult} & TAB & Financial & income $>$ 50K & 48 842 & \cite{liang2020think} \cite{abay2020mitigating} \cite{zhang2020fairfl} \cite{du2021fairness} \cite{padala2021federated} \cite{chu2021fedfair} \cite{cui2021addressing} \cite{rodriguez2021enforcing} \cite{mehrabi2022towards} \cite{zhang2022unified} \cite{papadaki2022minimax} \cite{nafea2022proportional} \cite{liu2022achieving} \cite{zeng2022improving} \cite{carey2022robust} \cite{touat2023towards} \cite{wang2023mitigating} \cite{hamman2023demystifying} \cite{salazar2023fair} \cite{ezzeldin2023fairfed} \cite{benarba2023empirical} \cite{gao2023effl} \cite{kyllo2023inflorescence} \cite{sun2023toward} \cite{sun2023toward} \cite{meerza2024glocalfair} \cite{zhang2024unified} \cite{yin2024distribution} \cite{salazar2024unveiling} \cite{badar2024fairtrade} \cite{su2024multi} \cite{chen4601101enforcing} & 67\% \\ \midrule
COMPAS \cite{compas} & TAB & Criminology & is rearrested & 7 214 & \cite{abay2020mitigating} \cite{zhang2020fairfl} \cite{chu2021fedfair} \cite{liu2022achieving} \cite{hu2022fair} \cite{zeng2022improving} \cite{carey2022robust} \cite{wang2023mitigating} \cite{salazar2023fair} \cite{ezzeldin2023fairfed} \cite{yin2024distribution} \cite{su2024multi} & 25\% \\ \midrule
CelebA \cite{celeba} & IMG & Faces & diverse features & 202 599 & \cite{kanaparthy2022fair} \cite{yang2022towards} \cite{hu2022fair} \cite{wang2023mitigating} \cite{chang2023bias} \cite{meerza2024glocalfair} \cite{djebrouni2024bias} \cite{dunda2024fairnessaware} & 17\% \\ \midrule
Dutch Census \cite{dutch} & TAB & Financial & occupation is high-level & 189 725 & \cite{du2021fairness} \cite{padala2021federated} \cite{salazar2023fair} \cite{zhang2024unified} \cite{benarba2023empirical} \cite{djebrouni2024bias} & 13\% \\ \midrule
Fashion-MNIST \cite{fashion-mnist} & IMG & Image Classification & clothing category & 70 000 & \cite{rodriguez2021enforcing} \cite{zhang2022unified} \cite{papadaki2022minimax} \cite{selialia2023mitigating} \cite{salazar2024unveiling} & 10\%\\ \midrule
Bank Marketing \cite{bank} & TAB & Financial & subscribe to the term deposit & 45 211 & \cite{padala2021federated} \cite{zeng2022improving} \cite{badar2024fairtrade} \cite{su2024multi} \cite{gu2022privacy} & 10\% \\ \midrule
KDD \cite{kdd} & TAB & Financial & income $>$ 50K & 299 285 & \cite{amiri2022impact} \cite{benarba2023empirical} \cite{djebrouni2024bias} \cite{badar2024fairtrade} & 8\% \\ \midrule
ACS-Employment \cite{acs-datasets} & TAB & Financial & is employed & 2 320 013 & \cite{papadaki2022minimax} \cite{hu2022fair} \cite{yang2023enhancing} \cite{chang2023bias} & 8\% \\ \midrule
eICU \cite{eicu} & TAB & Healthcare & different medical outcomes & 200 000 & \cite{cui2021addressing} \cite{gao2023effl} \cite{wang2024analyzing} & 6\% \\ \midrule
UTKFace \cite{utk-faces} & IMG & Faces & diverse features & 22 812 & \cite{meerza2024glocalfair} \cite{dunda2024fairnessaware} \cite{kanaparthy2022fair} & 6\% \\ \midrule
MEPS \cite{meps} & TAB & Healthcare & utilization of medical facilities & 35 428 & \cite{touat2023towards} \cite{benarba2023empirical} \cite{djebrouni2024bias} & 6\% \\ \midrule
ACSIncome \cite{acs-datasets} & TAB & Financial & income $>$ 50K & 1 599 229 & \cite{yang2023enhancing} \cite{yin2024distribution} \cite{chang2023bias} & 6\%  \\ \midrule
MNIST \cite{mnist} & IMG & Image Classification & digit recognition & 70 000 & \cite{selialia2023mitigating} \cite{salazar2024unveiling} & 4\% \\ \midrule
CIFAR-10 \cite{cifar10} & IMG & Image Classification & tiny image classification & 60 000 & \cite{papadaki2022minimax} \cite{selialia2023mitigating} & 4\% \\ \midrule
Credit Card Default \cite{credit-card-default} & TAB & Financial & default prediction & 30 000 & \cite{gu2022privacy} \cite{badar2024fairtrade} & 4\% \\ \midrule
Law \cite{law} & TAB & Educational & passes the bar exam & 20 798 & \cite{salazar2023fair} \cite{badar2024fairtrade} & 4\% \\ \midrule
ML-1M \cite{ml1m} & TAB & Entertainment & movie rating & 1 000 209 & \cite{pentyala2022privfairfl} \cite{agrawal2024no} & 4\%  \\ \midrule
Drug \cite{drug} & TAB & Healthcare & abuses volatile substance & 1 885 & \cite{chu2021fedfair} & 2\%  \\ \midrule
Heritage Health \cite{heritage-health} & TAB & Healthcare & Charleson Index (survival indicator) & 113 000 & \cite{mehrabi2022towards} & 2\%  \\ \midrule
Digits-five \cite{digit5} & IMG & Image Classification & digit recognition & 136 000 & \cite{zhang2022unified} & 2\%  \\ \midrule
dSprites \cite{dsprites} & IMG & Image Classification & shape & 737 280 & \cite{yang2022towards} & 2\%  \\ \midrule
Crime \cite{crime} & TAB & Criminology & violent crimes & 1 994 & \cite{liu2022achieving} & 2\%  \\ \midrule
ADS \cite{ads} & TAB & Advertisement & is interested & 36 000 & \cite{pentyala2022privfairfl} & 2\%  \\ \midrule
ML-100K \cite{ml1m} & TAB & Entertainment & movie rating & 100 000 & \cite{agrawal2024no} & 2\%  \\ \midrule
Amazon-Movies \cite{amazon-movies} & TAB & Entertainment & movie rating & 484 141 & \cite{agrawal2024no} & 2\%  \\ \midrule
ACSPublic-Coverage \cite{acs-datasets} & TAB & Financial & covered by public health insurance & 1 127 446 & \cite{chang2023bias} & 2\%  \\ \midrule
Synthea \cite{synthea} & TAB & Healthcare & mortality prediction & 1 000 000 & \cite{poulain2023improving} & 2\%  \\ \midrule
MIMIC-III \cite{mimic-iii} & TAB & Healthcare & mortality prediction & 53 423 & \cite{poulain2023improving} & 2\%  \\ \midrule
MIMIC-IV \cite{mimic-iv} & TAB & Healthcare & mortality prediction & 69 619 & \cite{wang2024analyzing} & 2\%  \\ \midrule
CAER-S \cite{caers} & VID & Emotion Recognition & emotion & 70 000 & \cite{selialia2023mitigating} & 2\%  \\ \midrule
Prostate Cancer \cite{prostate} & TAB & Healthcare & tumor type & 287 237 & \cite{zhang2024unified} & 2\%  \\ \midrule
Fetal State \cite{fetal-state} & TAB & Healthcare & cardio-vascular disease & 2 123 & \cite{zhang2024unified} & 2\%  \\ \midrule
COVID-19 \cite{covid} & TAB & Healthcare & mortality prediction & 6 882 & \cite{zhang2024unified} & 2\%  \\ \midrule
Support \cite{support} & TAB & Healthcare & mortality prediction & 1 000 & \cite{zhang2024unified} & 2\%  \\ \midrule
ARS \cite{ars} & TAB & Healthcare & activity recognition & 75 128 & \cite{djebrouni2024bias} & 2\%  \\ \midrule
MobiAct \cite{mobiact} & TAB & Healthcare & activity recognition & 16 756 325 & \cite{djebrouni2024bias} & 2\%  \\ \midrule
IPUMS \cite{ipums} & TAB & Financial & income $>$ 25K & 49 531 & \cite{chen4601101enforcing} & 2\%  \\ \midrule
Acute Inflammations \cite{acute-inflamations} & TAB & Healthcare & diagnosis & 120 & \cite{liang2023architectural} & 2\%  \\ \midrule
Synthetic & N.A. & N.A. & N.A. & N.A. & \cite{cui2021addressing} \cite{papadaki2022minimax} \cite{zeng2022improving} \cite{zeng2023federated} \cite{gao2023effl} & 10\% \\
\bottomrule
\end{longtable}
\end{small} 

Additionally, given the limited number of instances in some of these datasets, researchers have started using other larger datasets commonly used in fair machine learning, including MNIST \cite{mnist}, Fashion-MNIST \cite{fashion-mnist}, CIFAR-10 \cite{cifar10}, and CelebA \cite{celeba}. In some cases, they modify these datasets to create synthetic sensitive attributes. For example, \cite{salazar2024unveiling} modifies the MNIST dataset to introduce a sensitive attribute, $S$, with two groups ($S=1$ and $S=0$), representing distinct image characteristics. For $S=0$ images, they invert the background and digit colors compared to standard MNIST images ($S=1$).

Furthermore, some works have a particular focus on specific domains, such as healthcare. Poulain \textit{et al.} \cite{poulain2023improving} focus on healthcare applications using two datasets: 1) the Synthea dataset \cite{synthea}, a public synthetic EHR simulation program, and 2) MIMIC-III \cite{mimic-iii}, a real-world EHR dataset of ICU patients. Liang \textit{et al.} \cite{liang2023architectural} also address healthcare, combining it with blockchain technology. They propose a blockchain decentralized FL platform that improves fairness in predictive models within the healthcare domain while preserving privacy, using a dataset about inflammations of the bladder to predict acute inflammations \cite{acute-inflamations}. Zhang \textit{et al.} \cite{zhang2024unified} present a framework for achieving fairness in FL within healthcare institutions, conducting experiments on four medical datasets: 1) prostate cancer datasets from the US, 2) a fetal state dataset of cardiotocography, 3) a COVID-19 dataset of Brazilian patients, and 4) a support dataset of seriously ill hospitalized adults. Wang \textit{et al.} \cite{wang2024analyzing} assess the impact of personalized FL on group fairness in the healthcare domain through empirical analysis using two prominent real-world Electronic Health Records (EHR) datasets, namely eICU \cite{eicu} and MIMIC-IV \cite{mimic-iv}.

In terms of data types, the datasets used span various formats, with 30 being tabular, seven being image-based, and one being a video dataset. This diversity highlights the broad applicability of FL but also indicates a need for fairness solutions that extend beyond the predominantly tabular datasets to address the unique challenges posed by image and video data. Notably missing are text-based datasets, which are relevant for natural language processing tasks and would benefit from fairness interventions in FL.

Lastly, several works create synthetic datasets to explore group fairness in FL. For instance, Gao \textit{et al.} \cite{gao2023effl} generate a synthetic dataset with a protected attribute and general attributes following specific distributions, and labels are generated based on a defined mathematical relationship among these attributes. This approach allows researchers to systematically study the impact of different algorithms on fairness and model performance in a controlled environment.

To conclude, research on group fairness in FL has predominantly used traditional tabular datasets. To advance the field, it is important to explore a wider variety of data types and applications to better assess the effectiveness of fairness interventions in real-world scenarios. Expanding beyond these conventional datasets will help ensure that fairness solutions are applicable across diverse and practical contexts.

\section{Ethical, Legal, and Policy Considerations}\label{section:ethical-legal-policy}

As FL continues to gain attention in real-world applications, the ethical, legal, and policy dimensions surrounding its implementation become increasingly important. Ensuring that FL systems are fair, transparent, and aligned with societal values is important, particularly in sensitive domains such as healthcare, finance, and law enforcement, where AI-based decisions can have profound societal impacts. This section discusses the ethical, legal, and policy challenges inherent in deploying FL systems, focusing on fairness in FL, compliance with relevant laws and regulations, and the development of policy frameworks to govern their use in diverse contexts.

\subsection{Ethical Considerations}

Ethical machine learning design is essential to ensure that it respects societal values and promotes fairness, accountability, and transparency \cite{nakamura2024ethics}. Ethical guidelines, such as those set by the European Commission's High-Level Expert Group on Artificial Intelligence \cite{ec_ai_ethics_2019}, emphasize fairness as a core principle, ensuring that AI systems treat all individuals equitably and do not perpetuate or exacerbate existing inequalities. By addressing biases, machine learning systems have the potential to increase societal fairness and ensure that all individuals have equal access to opportunities, such as education, healthcare, and financial services. 

Despite the importance of ethical considerations in machine learning, there remains a lack of research specifically focused on the ethical challenges in FL, where ensuring group fairness becomes even more complex due to its decentralized nature. Since each client operates autonomously, there is no straightforward mechanism to ensure that fairness guidelines are being followed or that the system as a whole is not inadvertently amplifying biases present in any specific client's data. This raises important ethical questions about the responsibilities of both the model developers and the participating clients in mitigating discrimination and ensuring equitable treatment for all users. Furthermore, ethical considerations in FL extend to issues of informed consent and the equitable distribution of benefits and harms. Clients may not always fully understand the implications of participating in a federated system, particularly with respect to how their data will be used and how their local model training could impact global model outcomes. It is essential to ensure that informed consent is obtained, and that participants are aware of how their data may contribute to potential societal outcomes.

Ultimately, to align FL with ethical principles, the system must be designed to balance the autonomy of clients with the need for oversight, ensuring that the system operates in a way that benefits all people fairly. Ethical considerations must go beyond compliance and focus on creating systems that respect human dignity, prevent harm, and promote justice across all stakeholders.

\subsection{Legal Considerations}

Legal perspectives influence how machine learning systems, including FL models, are developed and deployed \cite{nakamura2024ethics}. Many group fairness approaches, such as statistical parity, aim to improve the position of historically disadvantaged minority groups, mirroring mechanisms such as affirmative action that seek to correct systemic inequalities \cite{abu2020contextual}.

As machine learning technologies evolve rapidly, existing legal frameworks often struggle to keep pace, creating a need for updated regulations that address the unique challenges presented by these systems \cite{nakamura2024ethics}. Several key legislative initiatives aim to address these concerns:
\begin{itemize}
    \item The AI Act \cite{ai-act} proposed by the European Commission is one such initiative, which seeks to create a regulatory framework categorizing AI systems based on their risk levels. This act imposes stricter requirements on higher-risk applications, ensuring that algorithms used in sensitive areas such as employment, healthcare, and law enforcement comply with fairness standards.
    \item Similarly, the Algorithmic Accountability Act \cite{algorithmic_accountability_act_2019} in the United States aims to establish mechanisms for auditing systems for biases and ensuring transparency. This law proposes creating requirements for algorithmic transparency and regular reporting on potential biases to enhance the accountability of AI systems.
    \item The General Data Protection Regulation (GDPR) \cite{gdpr} in the European Union provides a legal framework for protecting personal data, with an emphasis on data minimization and the `right to explanation'. However, these regulations may conflict with fairness-aware strategies in FL, which may require additional metadata about client distributions. The decentralized nature of FL also complicates accountability, as biased outcomes may arise from localized training data, rather than model-wide discrimination.
\end{itemize}

In addition to national regulations, international cooperation is essential to address the global nature of technologies \cite{nakamura2024ethics}. International efforts, such as the OECD's AI Principles \cite{oecd_ai_principles}, aim to develop shared standards and best practices for ethical AI. These initiatives promote collaboration among nations to build common regulatory frameworks that address the challenges of cross-border AI deployments.

Despite these advancements, legal frameworks must carefully consider the particularities of FL. The decentralized nature of FL complicates issues such as ensuring compliance with fairness and data protection standards. Without centralized oversight, it becomes difficult to track and enforce fairness guidelines across all participants, especially when different jurisdictions may be in place. Thus, legal regulations must adapt to the unique characteristics of FL to effectively address challenges such as transparency, accountability, and bias mitigation.

\subsection{Policy Considerations}

Policy frameworks are needed for guiding the development and deployment of FL systems, ensuring that fairness and accountability are maintained throughout their lifecycle. Lucaj \textit{et al.} \cite{lucaj2023ai} propose several key policy recommendations for addressing group fairness:
\begin{itemize}
    \item Current regulations focus mainly on auditing model outcomes, which may not fully address fairness issues. Audits should cover the entire AI lifecycle, from design to deployment and maintenance, enabling the detection and correction of bias at each stage.
    \item Governments should support independent AI research to enhance the governance of these technologies. Policymakers must collaborate with technical experts to create effective regulations.
    \item There is currently no standardized process for certifying AI systems, and gaps in existing regulations remain. Clear accreditation guidelines and regulatory infrastructure are needed to build trust and accountability.
\end{itemize}

Despite these policy recommendations, enforcing fairness in FL presents significant challenges and no work has focused on these considerations. Different stakeholders involved in FL, including organizations and regulatory bodies, may have varying definitions of fairness and distinct legal requirements. For example, a bank in one jurisdiction may face stricter anti-discrimination regulations than a healthcare provider in another jurisdiction. This makes it difficult to create a uniform policy that meets the needs of all stakeholders while ensuring fairness and compliance.

Addressing these challenges requires a multi-faceted policy approach that balances the autonomy of individual clients with the need for regulatory oversight and accountability. Policymakers should consider the unique characteristics of FL when designing regulations, ensuring that fairness is not compromised by the decentralized nature of this technology.

\section{Future Directions}\label{section:future-directions}

\begin{figure}[h!]
    \centering
    \includegraphics[width=\linewidth]{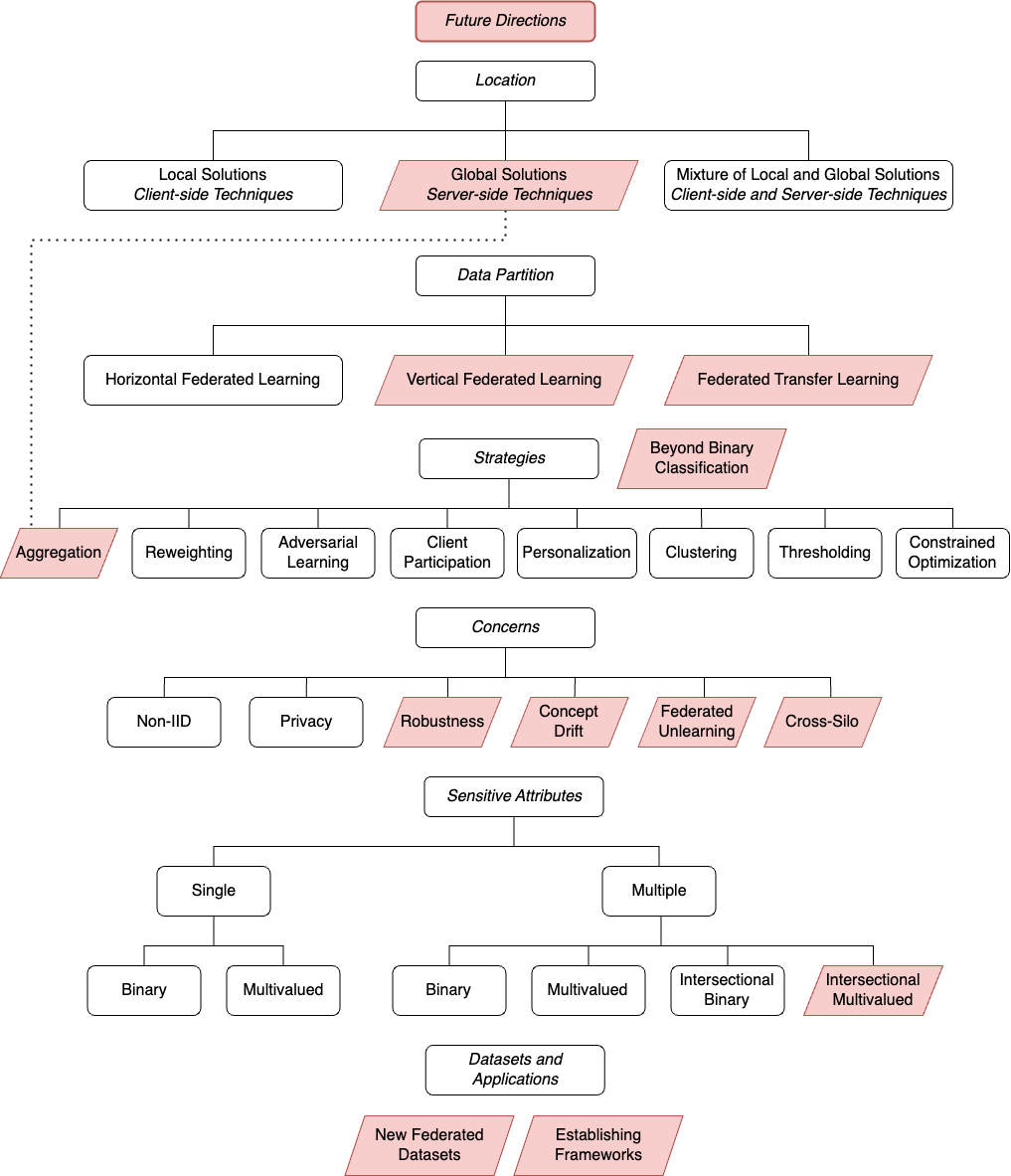}
    \caption{A new taxonomy of works on group fairness in Federated Learning, with future directions highlighted in red and represented by diamond shapes. The categories include location, data partition, strategies, concerns, sensitive attributes, and datasets and applications, providing a structured overview of the current state of research.}
    \label{fig:future-directions}
\end{figure}

As FL continues to evolve, addressing group fairness remains an important area of research. Table \ref{tab:works} presents a summary of the works discussed earlier in this paper. While significant progress has been made, numerous challenges and open questions persist. In this section, we highlight several possible future directions organized according to the new taxonomy of works on group fairness presented earlier in the paper, which includes the categories of: location, data partition, strategies, concerns, sensitive attributes, and datasets and applications. Figure \ref{fig:future-directions} illustrates this taxonomy, with key areas for future exploration highlighted in red and represented by diamond shapes.

\begin{small}
\begin{longtable}{p{0.75cm}|p{1.7cm}|p{3.6cm}|p{0.7cm}|p{0.6cm}|p{0.7cm}|p{1.35cm}|p{0.6cm}}
\caption{
Summary of works on group fairness in Federated Learning. \\
Location (LOC): L - local solutions, G - global solutions, H - hybrid solutions; \\
Data Partition (D.P.): HFL - horizontal FL, VFL - vertical FL; \\
Data: TAB - tabular, IMG - image, TXT - text; \\
Metrics: CD - conditioned on the decision, CO - conditioned on the outcome, PB - performance-based, UC - unconditional (e.g. SP), LB - loss-based; +DIF - difference, +RAT - ratio, +GS - group-specific, +AVG - average-based, +STD - standard deviation-based, +VAR - variance-based; \\
Sensitive Attributes (S.A.): SB - single binary, SV - single multivalued, MB - multiple binary; \\
N.A.: studies that do not propose specific methods but instead focus on conducting experiments related to group fairness.
}
\label{tab:works} \\
\toprule
\textbf{Work} & \textbf{Method} & \textbf{Focus} & \textbf{LOC} & \textbf{D.P.} & \textbf{Data} & \textbf{Metrics} & \textbf{S.A.} \\
\toprule
\endfirsthead
\toprule
Work & Method & Focus & LOC & D.P. & Data & Metrics & S.A. \\
\toprule
\endhead
\cite{liang2020think} & LG-FEDAVG & fair representations & L & HFL & TAB & N.A. (adv. loss) & MB \\ \midrule
\cite{abay2020mitigating} & - & local regularizer; local reweighting; global reweighting & L; H & HFL & TAB & UC+DIF UC+RAT CO+DIF & SB \\ \midrule
\cite{zhang2020fairfl} & FairFL & reinforcement learning; client selection & H & HFL & TAB & PB+DIF CO+DIF & SV \\ \midrule
\cite{du2021fairness} & AgnosticFair & kernel reweighting & H & HFL & TAB & UC+DIF & SB \\ \midrule
\cite{padala2021federated} & FPFL & differential privacy & H & HFL & TAB & UC+DIF CO+DIF & SB \\ \midrule
\cite{chu2021fedfair} & FedFair & constraint optimization; fairness estimation & H & HFL & TAB & CO+DIF & SB \\ \midrule
\cite{cui2021addressing} & FCFL & constrained multi-objective optimization; performance distribution fairness and group fairness simultaneously & H & HFL & TAB & UC+DIF CO+DIF & SB \\ \midrule
\cite{rodriguez2021enforcing} & FPFL & differential multipliers; fairness constraints & H & HFL & TAB IMG & PB+DIF UC+DIF CO+DIF CD+DIF & SV \\ \midrule
\cite{kanaparthy2022fair} & - & heuristics for aggregation techniques & G & HFL & IMG & PB+DIF CO+DIF & SV \\ \midrule
\cite{mehrabi2022towards} & FedVal & aggregation technique; identifying uncooperative clients & G & HFL & TAB & UC+DIF CO+DIF & SB \\ \midrule
\cite{zhang2022unified} & FMDA-M & multiple types of fairness simultaneously & H & HFL & TAB & PB+STD & SV \\ \midrule 
\cite{yang2022towards} & FairSCAT & adversarial learning & H & HFL & IMG & UC+DIF CO+DIF & SB \\ \midrule
\cite{papadaki2022minimax} & FedMinMax & minimax group fairness & H & HFL & TAB IMG & PB-GS PB-AVG & SV \\ \midrule
\cite{nafea2022proportional} & IFFCA & clustered FL & H & HFL & TAB & N.A. (unsup. notion) & SB \\ \midrule
\cite{amiri2022impact} & N.A. & impact of non-IID data & N.A. & N.A. & TAB & CO+DIF & SB \\ \midrule
\cite{pentyala2022privfairfl} & PrivFairFL & thresholding; secure multiparty computation and differential privacy & H & HFL & TAB & UC+DIF CO+DIF & SB \\ \midrule
\cite{gu2022privacy} & N.A. & trade-off between privacy, accuracy, and group fairness using differential privacy & N.A. & N.A. & TAB & UC+DIF CO+DIF & SB \\ \midrule
\cite{liu2022achieving} & FairVFL & vertical FL & H & VFL & TAB & CO+DIF & SB \\ \midrule
\cite{hu2022fair} & PFFL & bounded group loss; convergence and fairness guarantees & H & HFL & TAB IMG & PB+GS CO+GS UC+DIF CO+DIF & SB \\ \midrule
\cite{zeng2022improving} & FedFB & reweighting each subgroup & H & HFL & TAB & UC+DIF & SV \\ \midrule
\cite{carey2022robust} & FHN & personalization; fairness heterogeneity & H & HFL & TAB & UC+DIF CO+DIF & SB \\ \midrule
\cite{touat2023towards} & - & fairness and robustness & H & HFL & TAB & UC+DIF & SB \\ \midrule
\cite{wang2023mitigating} & FedGFT & local and global fairness & H & HFL & TAB IMG & UC+DIF CO+DIF CD+DIF & SB \\ \midrule 
\cite{poulain2023improving} & - & application: healthcare & H & HFL & TAB & PB+STD CD+STD CD+GS & SV \\ \midrule
\cite{hamman2023demystifying} & N.A. & local and global fairness trade-off & N.A. & N.A. & TAB & N.A. (mutual information) & SB \\ \midrule 
\cite{zeng2023federated} & LFT & local fair training & L & HFL & TAB & UC+DIF & SB \\ \midrule
\cite{salazar2023fair} & FAIR-FATE & fair aggregation using momentum techniques & G & HFL & TAB & UC+RAT CO+RAT & SB \\ \midrule
\cite{ezzeldin2023fairfed} & FairFed & client reweighting & H & HFL & TAB & CO+DIF & SB \\ \midrule
\cite{benarba2023empirical} & N.A. & impact of data size and heterogeneity & N.A. & N.A. & TAB & UC+DIF CO+DIF & SB \\ \midrule
\cite{chang2023bias} & N.A. & analysis of bias propagation & N.A. & N.A. & TAB & UC+DIF CO+DIF & SB \\ \midrule
\cite{selialia2023mitigating} & MWR & multiplicative weight update with regularization & H & HFL & IMG VID & PB+GS PB+AVG PB+VAR & SM \\ \midrule
\cite{yang2023enhancing} & N.A. & personalization & N.A. & N.A. & TAB & UC+DIF & SB \\ \midrule
\cite{gao2023effl} & EFFL & performance distribution fairness and group fairness simultaneously & H & HFL & TAB & PB+STD CO+STD & SB \\ \midrule 
\cite{kyllo2023inflorescence} & N.A. & impact of clustered on fairness & N.A. & N.A. & TAB & PB+RAT DI+RAT CO+RAT CD+RAT & SB \\ \midrule
\cite{liang2023architectural} & N.A. & application: healthcare & N.A. & N.A. & TAB & UC+RAT CO+DIF & SB \\ \midrule
\cite{sun2023toward} & FedLDP & trade-off between privacy, fairness and utility & H & HFL & TAB & UC+DIF CO+DIF & SB \\ \midrule 
\cite{meerza2024glocalfair} & GLocalFair & local and global fairness; constrained optimization; clustering & H & HFL & TAB IMG & UC+DIF CO+DIF & MB \\ \midrule 
\cite{djebrouni2024bias} & Astral & fair aggregation using a differential evolution algorithm & G & HFL & TAB IMG & PB+DIF UC+DIF CO+DIF & MB \\ \midrule
\cite{zhang2024unified} & FedUFO & application: healthcare & H & HFL & TAB & PB+STD PB+GS CO+DIF CO+GS & SB \\ \midrule
\cite{han2024fair} & FFL-OppoGAN & opposite generative adversarial networks; group fairness and performance distribution fairness simultaneously & H & HFL & TAB & UC+DIF & SB \\ \midrule 
\cite{dunda2024fairnessaware} & FFALM & constraint optimization; augmented Lagrangian method & H & HFL & IMG & UC+DIF CO+DIF & SB \\ \midrule 
\cite{yin2024distribution} & FedFaiREE & thresholding; distribution-free fair learning & H & HFL & TAB & CO+DIF & SB \\ \midrule 
\cite{salazar2024unveiling} & FairFedDrift & group-specific distributed concept drift; clustering & H & HFL & TAB IMG & PB+RAT CO+RAT CD+RAT & SB \\ \midrule 
\cite{wang2024analyzing} & N.A. & applications: healthcare; impact of personalization & N.A. & N.A. & TAB & UC+DIF CO+DIF & SV \\ \midrule 
\cite{badar2024fairtrade} & FairTrade & trade-off between balanced accuracy and fairness; multi-objective optimization & H & HFL & TAB & UC+DIF & SB \\ \midrule 
\cite{su2024multi} & mFairFL & group fairness and performance distribution fairness simultaneously; minimax constraint; gradient conflict detection & H & HFL & TAB & PB+DIF UC+DIF CO+DIF & SV \\ \midrule 
\cite{chen4601101enforcing} & DFLT; PGFD & local and global fairness; privacy constraints & H & HFL & TAB & UC+DIF CO+DIF & SB \\ \midrule
\cite{agrawal2024no} & F\textsuperscript{2}PGNN & personalized fair recommendation systems; graph neural networks & H & HFL & TAB & PB+DIF & SB \\

\bottomrule
\end{longtable}
\end{small} 

\subsection{Location}

\paragraph{Global Solutions} With respect to location in group fairness in FL, hybrid approaches are the most prevalent, as evidenced by Table \ref{tab:strategies}. These strategies, which involve collaboration between the server and clients, require significant computational resources and place additional responsibility on clients to comply with fairness requirements. Local approaches are less common in the literature due to their limitations in achieving global fairness, particularly when clients' data distributions are non-IID and fail to represent the global population adequately \cite{ezzeldin2023fairfed}. Global solutions, on the other hand, can address the limitations of both local and hybrid methods. Despite their potential, only four studies \cite{salazar2023fair, mehrabi2022towards, kanaparthy2022fair, djebrouni2024bias} have explored these global solutions, as discussed in Section \ref{sec:aggregation}, leaving room for improvement. For instance, none of these works has proposed efficient methods for creating the validation sets necessary for these aggregation algorithms. Future research could explore innovative methods to incentivize clients to contribute to the creation of more representative validation datasets. Additionally, existing studies could be expanded to analyze scenarios involving heterogeneous conditions, where some clients have varying fairness requirements.

\subsection{Data Partition}

\paragraph{Vertical Federated Learning} While most research in group fairness in FL has traditionally focused on HFL, the exploration of group fairness in VFL is equally relevant. In VFL, clients hold different subsets of features related to the same group of users, making it important to ensure fairness across these vertical partitions. As explained in Section \ref{sec:vertical}, group fairness in VFL introduces unique challenges compared to HFL. Nevertheless, only one work \cite{liu2022achieving} has focused on achieving group fairness in VFL. Future research could focus on exploring new fairness metrics that accommodate diverse feature sets and evaluating the impact of different vertical partitioning strategies on fairness outcomes.

\paragraph{Federated Transfer Learning} FTL, as discussed in Section \ref{sec:transfer}, is important in situations where traditional FL may not be feasible due to insufficient data in the target domain. Despite its significance, no current research explicitly addresses the intersection of FTL and group fairness, presenting an opportunity for future exploration. Ensuring fairness in FTL poses unique challenges that must be addressed, including: 1) the potential mismatch between the distributions of sensitive attributes in the source and target domains, which could lead to biased models, and 2) the complexity of defining and measuring fairness in FTL, as fairness metrics that work in the source domain may not be appropriate for the target domain. Future work should focus on developing strategies to ensure fairness in FTL while preserving privacy and maintaining model performance across diverse domains.

\subsection{Strategies}

\paragraph{Beyond Binary Classification} Most works on fairness in FL have focused on binary classification scenarios. However, there is a need to explore other types of learning tasks within FL, such as clustering, regression, recommendation systems, and others \cite{clustering-survey, vassoy2024consumer, agrawal2024no}. These scenarios present unique challenges that differ significantly from binary classification. For instance, in regression tasks, the fairness metrics and mitigation strategies need to account for continuous target variables, which introduces complexities in defining and measuring fairness within the context of FL. In addition, clustering in FL presents a particularly challenging scenario for group fairness. Different from supervised learning tasks where labels guide the training process, clustering is an unsupervised learning task where the objective is to group similar datapoints together without predefined labels. Ensuring that these clusters do not reinforce or exacerbate existing biases becomes difficult. The variability in data distributions across clients can lead to clusters that are not representative of the global population, potentially disadvantaging certain groups. The decentralized nature of FL adds another layer of complexity to these tasks. In clustering different clients might have data that forms distinct clusters locally, but these clusters might not align well with the global data distribution when aggregated. This misalignment can lead to unfair outcomes for certain groups. Addressing fairness in these diverse scenarios is important for developing comprehensive fairness-aware FL systems that can be applied across a wide range of applications.

\subsection{Concerns}

\paragraph{Robustness} Robustness refers to a system's ability to withstand adversarial attacks while maintaining consistent performance. As discussed in Section \ref{sec:robustness}, this challenge becomes more complex when group fairness is factored in, as it is important to differentiate between genuine minority group members and potential adversaries attempting to poison the model. Despite the importance of this issue, only one work \cite{touat2023towards} have explored the intersection of robustness and fairness in FL, highlighting the need for further research. For future work, Touat \textit{et al.} \cite{touat2023towards} propose using model inversion techniques combined with client data distribution analysis to identify Byzantine behavior. From there, selecting `honest and minority' clients could be based on how well their data contributes to better representation for minority groups. Another important point is that malicious clients can intentionally target minority groups. For instance, adversaries may manipulate their updates to disproportionately harm underrepresented groups, leading to unfair model outcomes even when overall model accuracy remains high. Hence, developing robust FL mechanisms that account for both adversarial threats and fairness constraints is an important research direction.

\paragraph{Concept Drift} Concept drift refers to the change in the statistical properties of the target variable that a model is trying to predict over time \cite{gama2014survey}. This phenomenon can significantly impact the performance and fairness of machine learning models, as they may become less accurate and more biased as the underlying data distribution changes. As mentioned in Section \ref{sec:concept}, ensuring group fairness in FL under concept drift presents several open challenges. A recent study by T. Salazar \textit{et al.} \cite{salazar2024unveiling} introduce the problem of group-specific distributed concept drift in FL. Group-specific distributed concept drift occurs when different clients in a FL setting experience distinct group-specific concept drifts. T. Salazar \textit{et al.} \cite{salazar2024unveiling} propose the FairFedDrift algorithm to address this issue. While this algorithm is an important step forward, it is computationally expensive and may not be practical for all FL scenarios. Future work should focus on developing more efficient and scalable algorithms to manage group-specific concept drift while maintaining fairness. This includes exploring adaptive methods that can dynamically adjust to changes in the data distribution, as well as investigating techniques to reduce the computational burden of this current solution. Ensuring that FL models can remain fair in the face of concept drift is an important area for ongoing research.

\paragraph{Federated Unlearning} Federated unlearning is an emerging technique that enables the removal of specific clients' contributions from a trained federated model without requiring full retraining from scratch \cite{lyu2022privacy}. This is particularly relevant for privacy concerns and regulatory compliance (e.g., GDPR's `right to be forgotten' \cite{gdpr}). However, machine unlearning can disproportionately impact certain sensitive groups, especially if a significant portion of data from underrepresented groups is removed. Although fairness in machine unlearning has started to gain importance in centralized settings \cite{zhang2024forgotten}, its implications for FL remain unexplored. As FL relies on aggregating updates from distributed clients, performing machine unlearning on specific clients while the remaining clients have more datapoints from privileged sensitive groups could amplify bias. This challenge becomes even more pronounced in fairness-aware FL algorithms, where unlearning may disrupt fairness constraints that were originally calibrated for the full client population. Hence, investigating the intersection of federated unlearning and fairness is an important research direction.

\paragraph{Cross-Silo} FL can be classified into two types based on the type of participating clients and the scale of model training: cross-device FL, where clients are typically mobile devices and the number of participants is large, and cross-silo FL, where clients are larger organizations and the number of participants is smaller \cite{huang2023promoting, huang2022cross}. While much of the existing literature on fairness in FL has focused on cross-device FL, cross-silo FL presents its own unique set of challenges. One such challenge is the issue of free-riding, where some clients contribute less to the FL process, such as submitting low-quality or minimal updates, but still benefit from the global model \cite{huang2023promoting}. In the case of fairness in cross-silo FL, free-riding may be more tolerable in certain situations, as it can allow institutions with more diverse data to still access the global model and its benefits, helping improve fairness in the model, even if their contributions are less frequent. Another important research direction in fairness in cross-silo FL is addressing the issue of business competitors among cross-silo clients. In these settings, competing organizations might be hesitant to fully participate in the FL process, as they may fear that sharing their data could undermine their own competitive advantage \cite{huang2023promoting}. However, this reluctance can lead to biases in the global model, as institutions with more diverse data could help improve the global model, benefiting all participants, even if they are competitors. Balancing collaboration and fairness in this context requires finding strategies that encourage cooperation and fairness while mitigating concerns about competitive disadvantage.

\subsection{Sensitive Attributes}

\paragraph{Addressing Intersectionality} Despite the increasing attention to fairness in FL, there have not been specific works that focus on intersectionality in group fairness within this domain. Intersectionality refers to the consideration of multiple sensitive attributes simultaneously, such as race, gender, and socioeconomic status, and how their interactions can lead to unique experiences of disadvantage or privilege \cite{wang2022towards, crenshaw1989demarginalizing}. Addressing intersectionality in FL is more complex than in centralized learning due to several intrinsic challenges. In a FL environment, data is distributed across multiple clients, each holding their own local datasets. These datasets may contain diverse distributions of sensitive attributes, which makes it challenging to ensure fairness across all intersections of these attributes. Different from centralized learning, where the entire dataset is available for analysis and fairness adjustments, FL must account for the decentralized nature of data and the limited visibility into the entire data distribution. Different clients may have different subgroups of interest, and the number of subgroups created by intersectionality can be very large. Consequently, certain subgroups might be under-represented or even absent in local datasets, while their representation can increase when considering the combined data from all clients. Achieving group fairness in such scenarios requires careful consideration of these diverse and distributed subgroups. Despite this, no work as focused on intersectionality in FL. Addressing these challenges is essential for building truly fair FL systems that consider the different realities of intersectional groups. 

\subsection{Datasets and Applications}

\paragraph{New Datasets} Most fairness datasets in machine learning do not reflect the distributed nature of FL, which often involves geographically or otherwise partitioned data across multiple clients. Common datasets such as the Adult \cite{adult} or the COMPAS \cite{compas} datasets that are typically used in centralized machine learning settings do not account for the decentralized, multi-client structure inherent to FL. In FL, each client's local dataset can exhibit different distributions, reflecting varying demographic or geographic characteristics. This non-IID data presents unique challenges for ensuring fairness across all clients. Existing fairness datasets do not adequately capture these characteristics, limiting their usefulness for developing and evaluating fairness-aware FL algorithms. Hence, there is a need to identify and curate real-world datasets that inherently possess the distributed nature of FL. Examples of such datasets could include healthcare data from multiple hospitals located in different regions, financial data from different financial institutions, and social media data from different platforms. These datasets would capture geographic diversity, varying demographic distributions, and diverse user behaviors, providing a realistic setting for studying group fairness under FL frameworks. Future work should focus on curating datasets that reflect the decentralized and diverse nature of real-world FL environments, enabling the development of fair FL algorithms that work in realistic scenarios.

\paragraph{Establishing Frameworks}

Currently, the lack of standardized frameworks in group fairness in FL research leads to significant variability in how algorithms are tested. This variability includes differences in datasets, data splits, number of participating clients, number of runs, number of communication rounds, and evaluation metrics used. As a result, it becomes difficult for researchers to compare the performance and fairness of different approaches effectively, making it challenging to draw meaningful and generalizable conclusions across studies. Establishing standardized frameworks for creating, testing, and validating fair FL algorithms is therefore important for advancing the field. Such frameworks would provide consistent guidelines for experimental setups, ensuring that algorithms are evaluated under comparable conditions. This consistency would not only facilitate more reliable comparisons between different approaches but also help identify best practices for achieving fairness in FL. Moreover, standardized frameworks would promote transparency and reproducibility in research, enabling the community to build on each other's work more effectively and accelerate the development of fair FL systems that can be deployed in real-world applications.

\section{Conclusions}\label{section:conclusions}

In this work, we introduced the first comprehensive survey focused on group fairness in FL. We discussed the unique challenges that arise due to the decentralized nature of FL, which complicates the implementation of fairness-aware algorithms. In addition, we proposed guidelines for identifying and benchmarking fairness in FL, providing a structured approach to evaluating different methods. Furthermore, we reviewed the current solutions that have been proposed to address these challenges, and examined the ethical, legal, and policy implications of fairness in FL. Finally, we highlighted future directions for research in this area.

The importance of ensuring group fairness in FL cannot be overstated. As FL continues to gain traction in various applications, from healthcare to finance, it is important to ensure that the models developed do not inadvertently perpetuate or amplify existing biases. Addressing group fairness in FL is not just a technical challenge but a societal imperative, as fair and equitable machine learning models can significantly impact people's lives.

We hope that this survey inspires researchers and practitioners to develop more fair FL systems that can be trusted to deliver equitable outcomes across all groups of the population. Furthermore, we aim to encourage further exploration of benchmarking methodologies, ethical and legal considerations, and practical deployment strategies to ensure fairness in real-world FL applications.

\backmatter

\bmhead{Acknowledgements}
This work was supported in part by the National Funds from the Fundação para a Ciência e a Tecnologia (FCT), I.P., under Project UIDB/00326/2025 and Project UIDP/00326/2025; in part by the Portuguese FCT Research under Grant 2021.05763.BD; and in part by the Portuguese Recovery and Resilience Plan (PRR) through the Center for Responsible AI under Project C645008882-00000055.

\section*{Declarations}

\noindent

{\setlength{\parindent}{0cm}\textbf{Conflict of interest} The authors report that there are no potential conflict of interest.} 
\\

{\setlength{\parindent}{0cm}\textbf{Data availability} The works and datasets analysed in this work are available at: \url{https://github.com/teresalazar13/Survey-Group-Fairness-in-Federated-Learning}.}
\\

{\setlength{\parindent}{0cm}\textbf{Author contributions} Teresa Salazar was responsible for the conceptualization, methodology, data collection, writing of the original draft, as well as reviewing and editing the manuscript. Helder Araujo and Alberto Cano were responsible for reviewing and editing the manuscript. Pedro Henriques Abreu provided supervision and was responsible for reviewing and editing the manuscript.}
\\

{\setlength{\parindent}{0cm}\textbf{Open Access} This article is licensed under a Creative Commons Attribution 4.0 International License, which permits use, sharing, adaptation, distribution and reproduction in any medium or format, as long as you give appropriate credit to the original author(s) and the source, provide a link to the Creative Commons licence, and indicate if changes were made. The images or other third party material in this article are included in the article's Creative Commons licence, unless indicated otherwise in a credit line to the material. If material is not included in the article's Creative Commons licence and your intended use is not permitted by statutory regulation or exceeds the permitted use, you will need to obtain permission directly from the copyright holder. To view a copy of this licence, visit \url{http://creativecommons.org/licenses/by/4.0/}.}

\bibliography{sn-bibliography.bib}

\end{document}